\documentclass[11pt]{article}

\usepackage[preprint]{acl}
\usepackage{subcaption}

\usepackage{tabularx} 
\usepackage{booktabs}
\usepackage{multirow}
\usepackage{amsmath}
\usepackage{amssymb}
\usepackage{url}
\usepackage{xcolor}
\usepackage{times}
\usepackage{latexsym}

\usepackage[T1]{fontenc}
\usepackage{enumitem}
\usepackage[utf8]{inputenc}

\usepackage{microtype}

\usepackage{inconsolata}

\usepackage{graphicx}
\definecolor{variantgray}{gray}{0.55}

\title{How Much Structure Do LLMs Need? Evaluating LLMs for Bibliometric Cluster Description}

\author{Abraham Camelo-Guerrero \\
School of Information Technology\\
  York University\\
  Toronto, Ontario M3J 1P3\\
  \texttt{acamelog@yorku.ca} \\\And
  Jairo Diaz-Rodriguez \\
Department of Mathematics and Statistics\\
  York University\\
  Toronto, Ontario M3J 1P3\\
  \texttt{jdiazrod@yorku.ca} \\}

\begin{document}
\maketitle
\begin{abstract}
Large language models (LLMs) can support scientific literature synthesis, but remain prone to hallucinated references, uneven coverage, and weakly grounded thematic organization. We evaluate whether bibliometric structure improves LLM-assisted synthesis by comparing six pipelines for generating cluster descriptions under different levels of evidence and structure. Using 100 published bibliometric analyses, we reconstruct Scopus corpora, extract human-written cluster descriptions, and assess outputs by human alignment, semantic coverage, clustering quality, graph quality, and reference grounding. Results show that LLMs produce descriptions semantically close to human-written ones, but are unreliable when asked to infer bibliometric structure from scratch. Performance improves when bibliometric algorithms define the clusters and the LLM interprets them. Overall, LLM-assisted bibliometric synthesis is most promising as a hybrid workflow in which algorithms provide auditable structure and LLMs generate readable descriptions.
\end{abstract}

\section{Introduction}

Large language models (LLMs) are increasingly used for scientific literature synthesis, including related-work generation, scientific summarization, and automatic literature review generation \citep{hu2014automatic,chen2021related,lu2020multi,kasanishi2023large,tang2025large}. Retrieval-augmented generation improves grounding in external corpora \citep{lewis2020retrieval,gao2023retrieval}, and recent systems target scientific synthesis directly \citep{asai2026openscholar}. However, generating a literature review from scratch remains difficult: LLMs may hallucinate references, provide uneven coverage, or impose an organization on the literature that is not supported by the scholarly record \citep{tang2025large}. A natural alternative is to separate organization from writing: first construct a structured map of the literature, then use the LLM to synthesize descriptions within that structure.

Bibliometric analysis provides such a structured approach. It uses publication and citation metadata to organize papers through relations such as bibliographic coupling, co-citation, and direct citation \citep{kessler1963bibliographic,small1973cocitation,boyack2010cocitation}. Widely used in management, information science, health sciences, environmental studies, education, and scientometrics, bibliometric science mapping provides an auditable way to identify research streams, intellectual foundations, influential works, and scholarly communities \citep{cobo2011science}. Recent work has begun to use LLMs around bibliometric analysis, mainly for auxiliary tasks such as search support, summarization, topic classification, and thematic mapping \citep{sarachuk2025still,he2025remote,keenan2026beyond}. However, these studies do not systematically evaluate how different levels of LLM responsibility affect the quality of the bibliometric workflow itself.  This makes bibliometric analysis a useful test case for LLM-assisted synthesis: if LLMs struggle to organize a literature from scratch, how much external structure should they receive?

\begin{figure*}[t]
    \centering
    \includegraphics[width=1\linewidth]{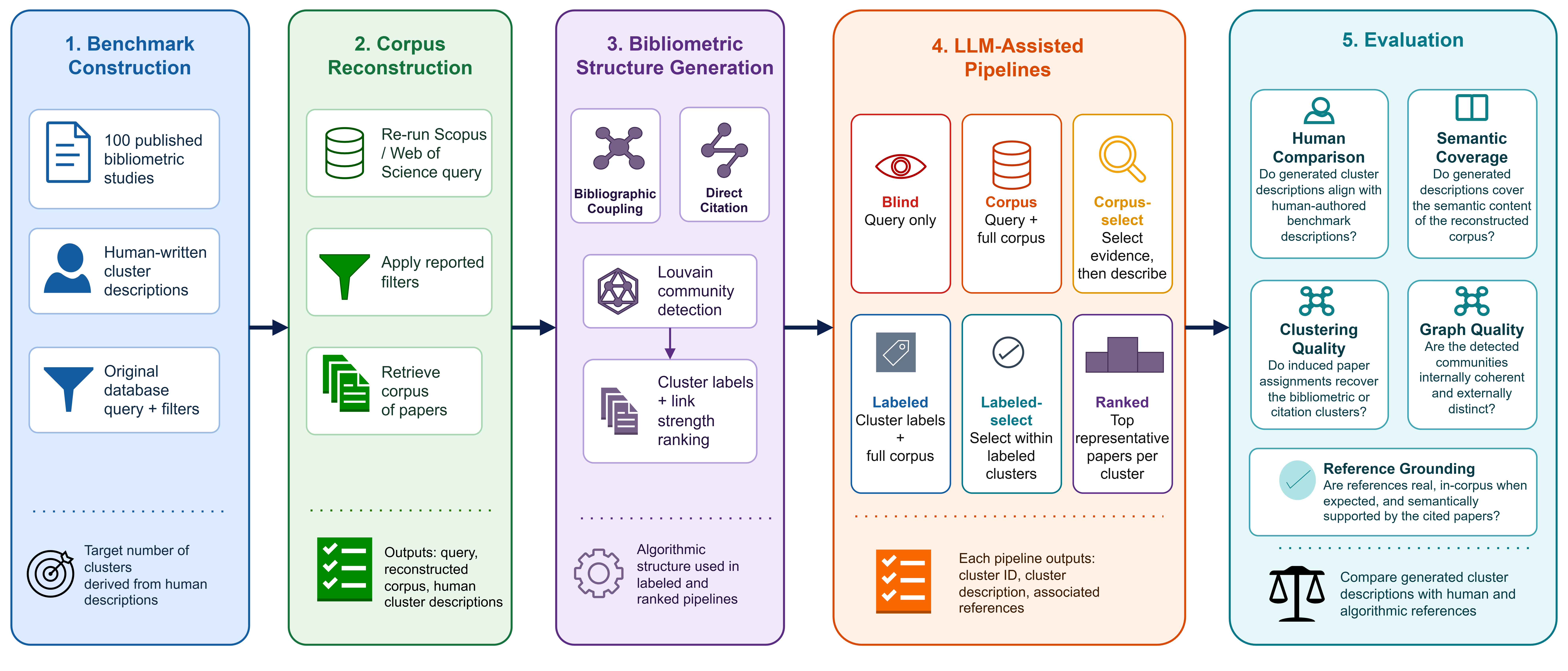}
    \caption{Workflow for evaluating LLM-assisted bibliometric synthesis.}
    \label{fig:workflow}
\end{figure*}

We address this question by evaluating six LLM-assisted pipelines for generating bibliometric cluster descriptions under different levels of evidence and structure (Figure~\ref{fig:workflow}). The pipelines range from \textit{Blind}, where the model receives only the search query, to structured settings such as \textit{Labeled} and \textit{Ranked}, where the model receives papers grouped by bibliographic coupling or direct citation clusters. We build the benchmark from 100 published bibliometric analyses by manually extracting author queries, reconstructing Scopus corpora, and collecting human-written cluster descriptions. We assess the outputs in terms of human alignment, semantic coverage, clustering quality, agreement with the underlying bibliometric graph, and reference grounding.

This design contributes to a broader shift in LLM-assisted literature review research from isolated single-task applications toward structured, multi-stage workflows. Recent work has examined LLMs for tasks such as screening, query generation, search, extraction, organization, and synthesis under human supervision within these multi-stage workflows \citep{nykvist2025testing,wang2025reassessing,ye2024hybrid,pei2025leveraging,silva2025enhancing}. Other studies emphasize structured outputs, including review tables, schemas, hierarchical maps, and extracted evidence, because these outputs are easier to inspect than free-form prose \citep{padmakumar2025intent,hsu2024chime,john2026extractable,jansen2025data}. Our study extends this direction by testing structure itself as an experimental condition.

Our results show that: (i) LLMs can generate cluster descriptions that are semantically close to human-written ones, but are unreliable when asked to infer bibliometric structure from scratch; (ii) when given appropriate bibliometric structure, LLM-generated descriptions can score higher than human descriptions on our corpus-level, clustering, and graph-based metrics, with the strongest performance occurring when bibliometric algorithms first define the clusters and the LLM is used to interpret them; and (iii) the best form of structure depends on the relation: bibliographic coupling benefits from full cluster context, while citation analysis can often be summarized from compact link-ranked evidence. (iv) Overall, LLM-based bibliometric analysis is most promising as a hybrid workflow in which algorithms provide auditable structure and LLMs translate that structure into readable descriptions.

\section{Related Work}
\label{sec:related-work}

NLP work on scientific literature synthesis includes related-work generation, scientific multi-document summarization, and automatic literature review generation \citep{hu2014automatic,chen2021related,lu2020multi,kasanishi2023large,tang2025large}. Recent retrieval-augmented systems further ground scientific generation in external corpora \citep{lewis2020retrieval,gao2023retrieval,asai2026openscholar}, while recent evaluations show that LLM-generated literature reviews still suffer from hallucinated references and uneven coverage \citep{tang2025large}. Our work differs by studying not full review generation, but the more specific task of generating cluster descriptions with the principles of bibliometric analysis \citep{cobo2011science}.

Our task connects topic labeling with bibliometric science mapping. Topic models and automatic labeling methods induce and verbalize latent themes from text \citep{blei2003latent,mei2007automatic,lau2011automatic,bhatia2016automatic}, and recent work uses LLMs to improve topic modeling and topic interpretability \citep{liu2025llm,yang2025neural}. Bibliometric methods instead cluster papers using relations such as bibliographic coupling, co-citation, and direct citation \citep{kessler1963bibliographic,small1973cocitation,boyack2010cocitation}, often with community detection algorithms such as Louvain \citep{blondel2008fast}. We use bibliometric clusters as an auditable scaffold for LLM generation, testing both whether LLMs can infer literature structure on their own and how much external structure is needed to improve cluster descriptions.

\section{Background: Bibliometric Analysis}

\subsection{Bibliometric Analysis as Science Mapping}

Bibliometric analysis is a structured approach for studying scientific literatures through metadata such as titles, abstracts, keywords, references, and citations. In science mapping, the goal is to reveal the organization of a research field: its main themes, intellectual foundations, influential works, and research communities \citep{cobo2011science}. Rather than producing a single linear summary, bibliometric analysis typically organizes a literature into clusters that can be interpreted as research streams or thematic areas.

A simplified bibliometric workflow consists of four stages \cite{donthu2021conduct}:

\[
\text{query}
\rightarrow
\text{corpus}
\rightarrow
\text{clustering}
\rightarrow
\begin{array}{c}
\text{cluster}\\
\text{descriptions}
\end{array}.
\]

\paragraph{Query.}
The query defines the scope of the analysis. It specifies which topic, keywords, time period, document types, or database fields are included in the search. Because all later stages depend on the retrieved papers, the query strongly shapes the resulting analysis. In published bibliometric reviews, the query is part of the methodological record and provides a reproducible entry point into the literature.

\paragraph{Corpus.}
Executing the query in a bibliographic database such as Scopus produces a corpus of papers. The corpus is the universe of documents to be organized. It provides textual evidence, such as titles and abstracts, as well as bibliographic evidence, such as references and citation links. Importantly, the corpus is not itself a clustering. It is a collection of papers that still needs to be structured.

\paragraph{Clustering.}
The clustering stage organizes the corpus into groups of related papers. Bibliometric clustering usually begins by defining relations among papers, then applying a clustering algorithm to those relations. Different relations capture different notions of scholarly relatedness. In this work, we focus on \emph{bibliographic coupling} and \emph{citation} analysis.

\emph{Bibliographic coupling} links two papers when they cite the same prior work \citep{kessler1963bibliographic}. Let \(R_i\) denote the set of references cited by paper \(p_i\). The bibliographic coupling weight between papers \(p_i\) and \(p_j\) is:

\[
w_{ij}^{BC} = |R_i \cap R_j|.
\]

If two papers share many references, they are likely to draw on similar intellectual foundations. Bibliographic coupling is therefore useful for identifying research fronts based on shared prior literature \citep{boyack2010cocitation}.

\emph{Citation} analysis links papers when one paper cites another \citep{garfield1955citation,price1965networks}. Let \(c_{ij}=1\) if paper \(p_i\) cites paper \(p_j\), and \(0\) otherwise. We use an undirected projection:

\[
w_{ij}^{CIT} = c_{ij} + c_{ji}.
\]

This relation captures direct scholarly influence within a retrieved corpus. Compared with bibliographic coupling, citation analysis emphasizes citation paths and intellectual lineage.

After constructing a bibliographic-coupling or citation-based representation, a clustering algorithm groups papers into communities. In our study, this stage is implemented using Louvain community detection \cite{yin2024quantitative}. The result is a paper-level clustering that specifies which papers belong together.

\paragraph{Cluster descriptions.}
A clustering identifies groups of related papers, but it does not explain what those groups mean. In traditional bibliometric reviews, human analysts interpret clusters by inspecting representative papers, central works, recurring terms, and citation patterns. They then write labels and descriptions that summarize each cluster's theme, scope, and relation to other clusters.

This final interpretive step is where LLMs may be especially useful. LLMs can synthesize text and produce fluent descriptions, but they may be unreliable when asked to infer an entire literature structure from a topic alone. A bibliometric workflow provides increasing structure: first a query, then a retrieved corpus, then an algorithmic clustering. This structure allows us to test whether LLMs are more reliable as autonomous cluster generators or as interpreters of clusters produced by standard bibliometric methods.

\section{Conceptual Framework}

We view LLM-assisted bibliometric analysis as a workflow-allocation problem: which stages should be handled by the LLM, and which should remain algorithmic? Our six pipelines form a ladder of increasing structure, from query-only generation, where the LLM must infer papers, themes, references, and clusters on its own, to structured workflows where a corpus is retrieved, a bibliographic or citation graph is built, clusters are assigned algorithmically, and the LLM only summarizes the resulting clusters. We hypothesize that LLMs are more reliable as interpreters of bibliometric structure than as unconstrained generators of literature maps, and test whether retrieved evidence and graph-induced clusters reduce hallucinated references, weak coverage, and unsupported thematic organization. Our goal is not to evaluate unconstrained topic discovery, but cluster-faithful bibliometric description generation: given or inferred evidence about a corpus, a system should produce descriptions whose induced interpretation preserves the bibliographic coupling or citation structure that bibliometric analysis is designed to expose.

\begin{table*}[t]
\centering
\small
\caption{The six pipelines form a continuum of decreasing LLM responsibility and increasing bibliometric structure.}
\label{tab:pipelines-conceptual}
\setlength{\tabcolsep}{3.5pt}
\renewcommand{\arraystretch}{1.08}
\begin{tabularx}{\textwidth}{@{}lXccX@{}}
\toprule
Pipeline & Input to LLM & Corpus & Cluster & LLM role \\
\midrule
Blind & Author query only & No & No & Blind generation of cluster descriptions \\
Corpus & Query and full Scopus corpus & Yes & No & Induce and describe clusters in one pass \\
Corpus-select & Query and full Scopus corpus, two steps & Yes & No & Select relevant papers, then synthesize descriptions \\
Labeled & Corpus pre-grouped by cluster & Yes & Yes & Describe given clusters in one pass \\
Labeled-select & Corpus pre-grouped by cluster, two steps & Yes & Yes & Select best papers within clusters, then synthesize \\
Ranked & Top-\(k\) papers ranked on each cluster & Yes & Yes & Summarize compact representative evidence \\
\bottomrule
\end{tabularx}

\end{table*}

\section{Methodology}

Our methodology follows the standard workflow of bibliometric analysis and uses it to define a controlled evaluation of LLM-assisted cluster description. We begin from published bibliometric review papers, extract the authors' original search queries and cluster descriptions, reconstruct the corresponding Scopus corpora, apply standard bibliometric clustering analysis, and then evaluate six LLM pipelines that receive different amounts of evidence and structure.

\subsection{Data: Source Bibliometric Review Papers}

The benchmark is constructed from 100 peer-reviewed published  bibliometric review papers, that we manually revised. Each source paper conducts a bibliographic coupling analysis of a research topic and reports a set of cluster descriptions. We use these papers because they provide naturally occurring examples of expert bibliometric analysis.

Let \(r_i\) denote the \(i\)-th source paper. From each source paper, we manually extract two artifacts. First, we extract the search query used by the authors, denoted by \(q_i\). Second, we extract the human-authored cluster descriptions reported in the paper, denoted by $H_i = \{h_{i1}, h_{i2}, \ldots, h_{iK_i}\},$ where \(K_i = |H_i|\) is the number of clusters reported by the source paper. The extracted query \(q_i\) defines the topic and retrieval scope of the original bibliometric analysis. The extracted human descriptions \(H_i\) provide the reference interpretation for the \emph{bibliographic coupling} analysis and determine the target number of clusters used in evaluation. Human-authored citation-cluster descriptions were not consistently available, so we leave human comparison for \emph{citation} analysis outside our scope.

\subsection{Benchmark Construction}

For each source paper \(r_i\), we reconstruct the bibliometric-analysis workflow in four stages:

\[
r_i
\rightarrow
q_i
\rightarrow
D_i
\rightarrow
Z_i
\rightarrow
Y_i.
\]

Here, \(q_i\) is the query extracted from the source paper, \(D_i\) is the Scopus corpus obtained by rerunning that query, \(Z_i\) is the clustering obtained by applying a standard bibliometric clustering analysis, and \(Y_i\) is the cluster description output generated by a pipeline.

\paragraph{Query extraction.}

For each bibliometric review paper \(r_i\), we manually identify the search query used by the original authors. This query may include keywords, Boolean operators, field restrictions, publication-year restrictions, or other search constraints reported in the paper. We denote the extracted query by \(q_i\).

\paragraph{Corpus reconstruction.}

We execute \(q_i\) in Scopus to obtain a corpus
$D_i = \{p_{i1}, p_{i2}, \ldots, p_{in_i}\}.$ Each paper \(p_{ij}\) contains a title, abstract, and citation metadata when available. 

\paragraph{Bibliometric clustering.}

For each reconstructed corpus \(D_i\), we apply a standard bibliometric clustering analysis. The scope of this paper considers two biblimetric analysis modes: bibliographic coupling or direct citation.
In the bibliographic-coupling mode, papers are related by shared references. In the direct-citation mode, papers are related by citation links among papers in \(D_i\). We construct the corresponding paper-relation representation and apply Louvain community detection to obtain a paper-level clustering
$Z_i = \{z_{i1}, z_{i2}, \ldots, z_{in_i}\},
$ where $z_{ij} \in \{1, \ldots, K_i\}
$ is the cluster assignment of paper \(p_{ij}\).

The number of clusters is fixed to \(K_i\), the number of human-authored cluster descriptions extracted from the source paper. We tune the Louvain resolution parameter only to match this target number of clusters. We do not tune the clustering to optimize agreement with the human descriptions or with any downstream evaluation metric.

\paragraph{Cluster description generation.}

Each evaluated pipeline produces a set of cluster descriptions

\[
Y_i =
\{(l_{i1}, d_{i1}), \ldots, (l_{iK_i}, d_{iK_i})\},
\]

where \(l_{ik}\) is the generated label for cluster \(k\), and \(d_{ik}\) is its generated natural-language description.

Different pipelines receive different objects from the reconstructed workflow. Some receive only the extracted query \(q_i\). Others receive the reconstructed corpus \(D_i\). The most structured pipelines receive the Louvain clustering \(Z_i\) and ask the LLM only to describe the resulting clusters.

\begin{figure*}[ht]
    \centering

    \begin{minipage}[t]{0.21\textwidth}
        \centering
        \includegraphics[width=\linewidth]{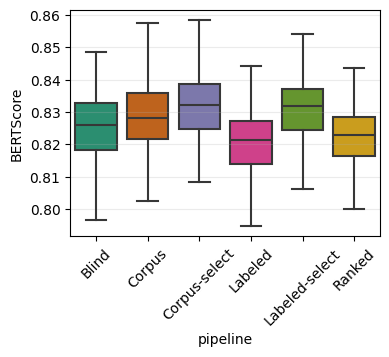}
        \caption{Pipeline Human comparison, measured with BERTScore.}
        \label{fig:BERTScore}
    \end{minipage}
    \hfill
    \begin{minipage}[t]{0.21\textwidth}
        \centering
        \includegraphics[width=\linewidth]{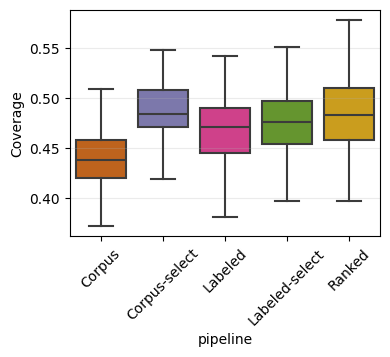}
        \caption{Reference-grounded coverage across pipelines.}
        \label{fig:reference-grounded}
    \end{minipage}
    \hfill
    \begin{minipage}[t]{0.54\textwidth}
        \centering
        \includegraphics[width=\linewidth]{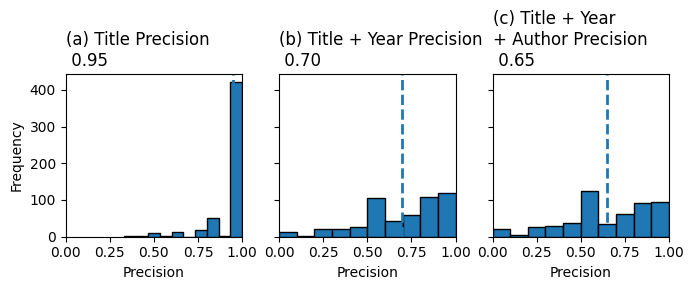}
        \caption{Reference validation for \textit{Blind} by matching criterion: (a) title, (b) title and year, and (c) title, year, and first author.}
        \label{fig:cluster-level-reference}
    \end{minipage}

\end{figure*}

\subsection{Pipeline Design}

The six pipelines differ in how much evidence and bibliometric structure are provided to the LLM. They form a progression from \textit{Blind}, where the model receives only the author query, to \textit{Ranked}, where the clusters have already been produced by a Louvain-based bibliometric or citation procedure and the model receives only a compact set of representative papers to summarize.  The pipeline ladder is designed to progressively reduce the structural responsibility assigned to the LLM. In the least structured setting, the model must infer the relevant themes and produce cluster descriptions from the query alone. In the most structured setting, both the cluster structure and the evidence selection are supplied by the bibliometric procedure, leaving the LLM primarily responsible for final verbalization. A detailed description of the pipelines and prompts is provided in Appendix~\ref{app:pipelines} and~\ref{app:prompt-schema}.

\paragraph{Blind. }

The \textit{Blind} pipeline receives only the extracted author query \(q_i\) and the target number of clusters \(K_i\). The LLM is asked to generate \(K_i\) cluster labels and descriptions directly from the query.
This is the least informed baseline. It must rely on its knowledge to infer the main research areas associated with the query. This pipeline approximates a naive LLM-based literature mapping prompt, except that the number of clusters is controlled.

\paragraph{Corpus. }

The \textit{Corpus} pipeline receives the extracted author query \(q_i\), the target number of clusters \(K_i\), and the full reconstructed Scopus corpus \(D_i\), represented by paper titles and abstracts. The LLM is asked to generate all cluster labels and descriptions in a single pass.
This pipeline tests whether access to the retrieved corpus is sufficient for the model to organize the literature. The LLM receives substantially more evidence than in \textit{Blind}, but it still does not receive an algorithmic clustering. It must both infer the cluster structure and describe the resulting clusters.

\paragraph{Corpus-select. }

The \textit{Corpus-select} pipeline receives the same inputs as \textit{Corpus}. However, it decomposes the task into two stages.
In the first stage, the LLM selects papers from \(D_i\) that are relevant or representative for each of the \(K_i\) clusters. In the second stage, the model synthesizes cluster labels and descriptions from the selected papers.
This pipeline tests whether explicit evidence selection improves corpus-grounded cluster analysis. The LLM still remains responsible for inducing the cluster structure, but it is encouraged to filter the corpus before producing the final descriptions.

\paragraph{Labeled. }

The \textit{Labeled} pipeline receives the corpus pre-grouped according to the Louvain clustering \(Z_i\). The LLM is given papers from \(D_i\) together with their algorithmic cluster assignments. The model is then asked to generate labels and descriptions for the given clusters in a single pass.
This pipeline provides strong structural information. Unlike \textit{Corpus} and \textit{Corpus-select}, the LLM no longer needs to decide which papers belong together. Its role shifts from cluster induction to cluster interpretation: it must explain the meaning of clusters produced by the standard bibliometric or citation clustering procedure.

\paragraph{Labeled-select}

The \textit{Labeled-select} pipeline also receives the corpus pre-grouped by the Louvain clustering \(Z_i\), but uses a two-stage procedure.
In the first stage, the LLM selects the most informative or representative papers within each predefined cluster. In the second stage, it synthesizes the final cluster labels and descriptions from the selected within-cluster evidence.
This pipeline tests whether evidence selection remains useful when the cluster structure is already given. Compared with \textit{Labeled}, the model receives the same algorithmic grouping, but is asked to identify the most useful evidence inside each cluster before writing the description.

\paragraph{Ranked}

The \textit{Ranked} pipeline also starts from the Louvain clustering \(Z_i\), but provides the LLM with only a compact subset of papers from each cluster. For each cluster, papers are ranked using link-based importance derived from the bibliometric clustering algorithm. The top 10 papers per cluster are then provided to the LLM, which generates the cluster labels and descriptions.
This pipeline tests whether a compact, high-signal subset of each algorithmic cluster is sufficient for accurate cluster description. The LLM receives less context than in \textit{Labeled} and \textit{Labeled-select}, but it also has less structural responsibility: the clusters have already been defined by Louvain, and the evidence shown to the model has already been ranked by the bibliometric procedure. The model's role is therefore limited to synthesizing descriptions from representative evidence, making \textit{Ranked} the most constrained pipeline in terms of LLM decision-making.

\begin{table*}[!t]
\centering
\caption{Main results for bibliographic coupling and citation analyses for all pipelines. For each metric, \emph{Rank} is the mean rank across benchmark instances, where lower is better; \emph{Med.} is the median value; and \emph{\%Win} is the percentage of instances won, where higher is better. Bold values mark the best result within each analysis block.}
\label{tab:main-results-combined}
\setlength{\tabcolsep}{3pt}
\renewcommand{\arraystretch}{0.95}
\resizebox{\textwidth}{!}{%
\footnotesize
\begin{tabular}{l*{12}{c}}
\toprule
\textbf{Pipeline}
& \multicolumn{3}{c}{{Semantic}}
& \multicolumn{6}{c}{{Cluster quality}}
& \multicolumn{3}{c}{{Graph quality}} \\
\cmidrule(lr){2-4}
\cmidrule(lr){5-10}
\cmidrule(lr){11-13}
& \multicolumn{3}{c}{\textbf{Coverage}}
& \multicolumn{3}{c}{\textbf{Silhouette}}
& \multicolumn{3}{c}{\textbf{ARI}}
& \multicolumn{3}{c}{\textbf{Modularity}} \\
\cmidrule(lr){2-4}
\cmidrule(lr){5-7}
\cmidrule(lr){8-10}
\cmidrule(lr){11-13}
& \textbf{Rank} & \textbf{Med.} & \textbf{\%Win}
& \textbf{Rank} & \textbf{Med.} & \textbf{\%Win}
& \textbf{Rank} & \textbf{Med.} & \textbf{\%Win}
& \textbf{Rank} & \textbf{Med.} & \textbf{\%Win} \\
\midrule
\multicolumn{13}{l}{\textit{Bibliographic coupling}} \\
\midrule
Blind
& 5.93 & 0.4081 & 0.00
& 4.71 & 0.0275 & 10.00
& 5.38 & 0.0534 & 4.00
& 5.515 & 0.0979 & 2.02 \\

Corpus
& 6.22 & 0.4084 & 0.00
& 3.57 & \textbf{0.0402} & 18.00
& 4.95 & 0.0762 & 0.00
& 3.955 & 0.1549 & 17.17 \\

Corpus-select
& 3.75 & 0.4294 & 5.00
& 3.60 & 0.0366 & 15.00
& 4.95 & 0.0766 & 5.00
& 3.944 & 0.1396 & 12.12 \\

Labeled
& \textbf{1.86} & \textbf{0.4474} & \textbf{45.00}
& \textbf{3.23} & 0.0398 & \textbf{21.00}
& \textbf{1.85} & \textbf{0.1828} & \textbf{51.00}
& 3.207 & 0.1693 & 23.23 \\

Labeled-select
& 3.95 & 0.4329 & 2.00
& 4.04 & 0.0350 & 10.00
& 2.93 & 0.1499 & 17.00
& 3.631 & 0.1721 & 14.14 \\

Ranked
& 2.46 & 0.4348 & 21.00
& 3.93 & 0.0376 & 15.00
& 2.50 & 0.1605 & 22.00
& \textbf{2.874} & \textbf{0.1830} & \textbf{29.29} \\

\emph{Human}
& \emph{3.83} & \emph{0.4316} & \emph{27.00}
& \emph{4.92} & \emph{0.0260} & \emph{11.00}
& \emph{5.44} & \emph{0.0457} & \emph{1.00}
& \emph{4.874} & \emph{0.0857} & \emph{7.07} \\

\midrule
\multicolumn{13}{l}{\textit{Citation}} \\
\midrule
Blind
& 5.70 & 0.4164 & 0.00
& 4.45 & 0.0273 & 15.00
& 5.300 & 0.1053 & 5.00
& 5.00 & 0.1841 & 0.00 \\

Corpus
& 5.10 & 0.4254 & 0.00
& 3.55 & 0.0379 & \textbf{20.00}
& 4.875 & 0.1329 & 0.00
& 4.15 & 0.2366 & 5.00 \\

Corpus-select
& 2.45 & 0.4384 & \textbf{35.00}
& 3.75 & 0.0367 & \textbf{20.00}
& 4.425 & 0.1550 & 0.00
& 3.40 & 0.2443 & 15.00 \\

Labeled
& \textbf{2.30} & \textbf{0.4444} & 25.00
& 3.05 & 0.0396 & 15.00
& \textbf{1.950} & 0.3246 & 40.00
& 2.75 & 0.2956 & 25.00 \\

Labeled-select
& 2.45 & 0.4391 & 30.00
& \textbf{3.00} & \textbf{0.0446} & \textbf{20.00}
& 2.500 & 0.2975 & 10.00
& 3.55 & 0.2943 & 5.00 \\

Ranked
& 3.00 & 0.4433 & 10.00
& 3.20 & 0.0404 & 10.00
& \textbf{1.950} & \textbf{0.3326} & \textbf{45.00}
& \textbf{2.15} & \textbf{0.3153} & \textbf{50.00} \\
\bottomrule
\end{tabular}%
}
\end{table*}

\section{Evaluation}
\label{sec:evaluation}

 We evaluate each generated cluster analysis along five dimensions that reflect the main goals of bibliometric analysis: alignment with human interpretations, semantic coverage of the corpus, recovery of cluster structure, agreement with the underlying bibliometric graph, and reference grounding. Full metric definitions are provided in Appendix~\ref{app:evaluation-details}. We complement this scalable structure-sensitive automatic metrics with a small-scale human evaluation reported in Appendix~\ref{app:human-evaluation}.

\paragraph{Human comparison.}
We use BERTScore F1 to measure semantic alignment between generated cluster descriptions and the human-authored descriptions reported in the source bibliometric reviews. Generated and human descriptions are matched one-to-one to maximize total semantic similarity. This captures whether the generated descriptions recover similar themes, even when they use different wording.

\paragraph{Semantic coverage.}
We measure coverage as the average maximum embedding similarity between each sentence in the reconstructed Scopus corpus and the generated cluster descriptions. This evaluates whether the pipeline describes the broader corpus rather than only a narrow subset of papers.

\paragraph{Clustering quality.}
We assign each paper to the generated cluster description with the highest embedding similarity and compare these induced assignments with the bibliometric or citation clusters using Adjusted Rand Index. We also report silhouette score in abstract embedding space. These metrics test whether the generated descriptions preserve the cluster structure intend to explain.

\paragraph{Graph-structural quality.}
We compute modularity using the induced paper assignments on the underlying bibliographic or citation graph. This evaluates whether the descriptions correspond to dense regions of the paper-relation network, rather than only matching the final Louvain labels.

\paragraph{Reference grounding.}
For \textit{Blind}, we validate generated references against OpenAlex using title, year, and first-author matching. For the other pipelines, we compute reference-grounded coverage between cited abstracts and generated descriptions. This tests whether references are valid and whether descriptions reflect the evidence they cite.

\section{Results}
\label{sec:results}

 Figure~\ref{fig:BERTScore} reports semantic alignment with the human-authored cluster descriptions, Figure~\ref{fig:reference-grounded} and  Figure~\ref{fig:cluster-level-reference} show reference-grounded results.
Table~\ref{tab:main-results-combined} summarizes the main results for bibliographic coupling and citation analyses across metrics.

\paragraph{Human comparison.}
Figure~\ref{fig:BERTScore} shows consistently high BERTScore values above 0.8 across all pipelines, indicating strong semantic overlap with human-written cluster descriptions. Table~\ref{tab:main-results-combined} shows a different pattern when the outputs are evaluated against corpus-level and graph-structural metrics. Human descriptions achieve reasonable coverage, but they rank closer to the middle on average and perform poorly on cluster and graph quality. This does not mean that the human descriptions are low quality in an absolute sense. Rather, it shows that descriptions that are semantically similar to human-written summaries do not necessarily induce paper assignments that recover the bibliometric or citation structure. Structured LLM pipelines often achieve higher values than the human descriptions on these structure-sensitive metrics, suggesting that they can complement human-written descriptions when the goal is to summarize corpus coverage and clustering structure in bibliometric analyses.

\paragraph{Structure improves performance.}
Table~\ref{tab:main-results-combined} also shows that LLMs are weak when they must infer bibliometric structure on their own. The \textit{Blind} pipeline performs poorly in both bibliographic coupling and citation settings, suggesting that query-only generation is not reliable for autonomous bibliometric mapping. Figure~\ref{fig:cluster-level-reference} further shows that this weakness extends to reference grounding. Under title-only matching, many \textit{Blind} references appear to correspond to real papers, but when the validation also requires the publication year and first author to match, precision drops. Thus, \textit{Blind} can often produce references that look plausible, but it is less reliable at generating fully accurate citations.

Providing the full Scopus corpus helps: \textit{Corpus} and \textit{Corpus-select} improve over \textit{Blind}. Figure~\ref{fig:reference-grounded} further shows that reference-grounded coverage is strongest for pipelines that rely on selected or representative evidence, especially \textit{Corpus-select} and \textit{Ranked}, while \textit{Corpus} performs lowest. This suggests that evidence selection improves local grounding.
However, Table~\ref{tab:main-results-combined} shows corpus-only pipelines still lag behind pipelines that include clustering structure on cluster and graph quality. Seeing the papers improves grounding, but it does not guarantee that the model recovers bibliographic or citation structure. A strong conclusion is that LLMs are better used as interpreters of bibliometric structure than as generators of that structure from scratch. \textit{Labeled}, \textit{Labeled-select}, and \textit{Ranked} perform best overall. This reflects the main design principle of the pipeline ladder: as more structure is provided to the model, the LLM has less freedom to invent or infer the organization of the field. In the structured pipelines, the bibliometric clustering determines which papers belong together, and the LLM is increasingly restricted to interpreting those clusters. The same pattern appears in both bibliographic coupling and citation analysis, but the best form of structure differs by mode.

\paragraph{Task-dependent structure. }The best amount of structure depends on the bibliometric relation being summarized. Bibliographic coupling groups papers through shared references, so its clusters often reflect broad intellectual backgrounds or research traditions. Because these themes may be distributed across both central and peripheral papers, \textit{Labeled} performs especially well: the model benefits from access to the full labeled cluster. Citation analysis, by contrast, is based on direct links among papers and often concentrates structure around influential or central works. In this setting, \textit{Ranked} performs especially well because top link-ranked papers can capture much of the cluster's lineage, influence pattern, or methodological core. Overall, broad shared-reference clusters benefit from richer labeled context, while citation-based clusters can often be described effectively from compact, central evidence.

\paragraph{Ablations. }We conduct four ablation studies to test the robustness of the pipeline comparisons (Appendix \ref{sec:ablations}). We ablate the generative model by comparing GPT-5.4 and Gemini-2.5-Flash, the embedding model used for evaluation by comparing \texttt{all-mpnet-base-v2} and \texttt{text-embedding-3-large}, the prompt formulation by removing or adding query and bibliographic-coupling context, and the amount of evidence by varying the number of references or top-ranked papers provided per cluster. Across these ablations, the main patterns described above remain stable. Prompt and evidence-size choices can shift which structured pipeline performs best, but they do not change the broader conclusion that evidence structure matters more than prompt wording input parameters.

\section{Conclusions}

In this paper, we evaluated how LLMs support bibliometric cluster analysis by comparing six pipelines with increasing levels of evidence and structure. Using published bibliometric reviews, we extracted author queries and human cluster descriptions, reconstructed Scopus corpora, and generated cluster descriptions under different workflow designs. Our results show that LLMs can produce descriptions that are semantically similar to human-written ones and, when given bibliometric structure, can achieve higher values than human descriptions on corpus-level, clustering, and graph-based metrics. However, they remain weak at inferring bibliometric structure from scratch, even with access to the full corpus. They work best when bibliometric algorithms first define the clusters and the LLM is used to interpret them.

The best pipeline also depends on the bibliometric relation. Bibliographic coupling benefits from full cluster context, while citation analysis can often be summarized from compact link-ranked evidence. This suggests that structure helps most when it constrains parts of the task where LLMs are unreliable, but the right amount of structure depends on the reasoning required. Overall, LLM-based bibliometric analysis is most promising as a hybrid workflow: algorithms provide auditable structure, and LLMs translate that structure into readable cluster descriptions.

This finding reflects a broader pattern in LLM system design. Across retrieval-augmented generation, tool use, code generation, and data analysis, LLMs are most reliable when external systems provide structure and the model performs synthesis, explanation, or translation. In bibliometric analysis, algorithms should handle precise structural tasks such as clustering papers by citations or shared references, while LLMs should handle the interpretive task of writing coherent descriptions.

\paragraph{Limitation. }These results should be interpreted as evidence of structural fidelity, not as a general comparison between LLMs and human experts. Human descriptions were written for scholarly interpretation, not to optimize embedding-based coverage, ARI, or modularity. Thus, when structured LLM pipelines score higher on these metrics, it means they better preserve the reconstructed bibliometric or citation structure used in our evaluation, not that they offer greater domain insight, nuance, or scholarly value.

\section*{Acknowledgments}
This work was supported by the Natural Sciences and Engineering Research Council of Canada (NSERC) under grant DGECR-2022-04531.

\bibliography{custom}

@article{cobo2011science,
  title   = {Science Mapping Software Tools: Review, Analysis, and Cooperative Study Among Tools},
  author  = {Cobo, Manuel J. and L{\'o}pez-Herrera, Antonio Gabriel and Herrera-Viedma, Enrique and Herrera, Francisco},
  journal = {Journal of the American Society for Information Science and Technology},
  volume  = {62},
  number  = {7},
  pages   = {1382--1402},
  year    = {2011},
}

@article{kessler1963bibliographic,
  title   = {Bibliographic Coupling Between Scientific Papers},
  author  = {Kessler, Maxwell Mirton},
  journal = {American Documentation},
  volume  = {14},
  number  = {1},
  pages   = {10--25},
  year    = {1963},
}

@article{newman2004modularity,
  title   = {Fast Algorithm for Detecting Community Structure in Networks},
  author  = {Newman, Mark E. J.},
  journal = {Physical Review E},
  volume  = {69},
  number  = {6},
  pages   = {066133},
  year    = {2004},
}

@article{price1965networks,
  title   = {Networks of Scientific Papers: The Pattern of Bibliographic References Indicates the Nature of the Scientific Research Front},
  author  = {Price, Derek J. de Solla},
  journal = {Science},
  volume  = {149},
  number  = {3683},
  pages   = {510--515},
  year    = {1965}
}

@article{small1973cocitation,
  title={Co-citation in the scientific literature: A new measure of the relationship between two documents},
  author={Small, Henry},
  journal={Journal of the American Society for information Science},
  volume={24},
  number={4},
  pages={265--269},
  year={1973},
  publisher={Wiley Online Library}
}

@article{boyack2010cocitation,
  title   = {Co-citation Analysis, Bibliographic Coupling, and Direct Citation: Which Citation Approach Represents the Research Front Most Accurately?},
  author  = {Boyack, Kevin W. and Klavans, Richard},
  journal = {Journal of the American Society for Information Science and Technology},
  volume  = {61},
  number  = {12},
  pages   = {2389--2404},
  year    = {2010}
}

@article{rousseeuw1987silhouettes,
  title   = {Silhouettes: A Graphical Aid to the Interpretation and Validation of Cluster Analysis},
  author  = {Rousseeuw, Peter J.},
  journal = {Journal of Computational and Applied Mathematics},
  volume  = {20},
  pages   = {53--65},
  year    = {1987}
}

@article{hubert1985comparing,
  title   = {Comparing Partitions},
  author  = {Hubert, Lawrence and Arabie, Phipps},
  journal = {Journal of Classification},
  volume  = {2},
  number  = {1},
  pages   = {193--218},
  year    = {1985}
}

@article{garfield1955citation,
  title   = {Citation Indexes for Science: A New Dimension in Documentation through Association of Ideas},
  author  = {Garfield, Eugene},
  journal = {Science},
  volume  = {122},
  number  = {3159},
  pages   = {108--111},
  year    = {1955}
}

@inproceedings{hu2014automatic,
  title     = {Automatic Generation of Related Work Sections in Scientific Papers: An Optimization Approach},
  author    = {Hu, Yue and Wan, Xiaojun},
  booktitle = {Proceedings of the 2014 Conference on Empirical Methods in Natural Language Processing (EMNLP)},
  pages     = {1624--1633},
  year      = {2014}
}

@inproceedings{chen2021related,
  title     = {Capturing Relations Between Scientific Papers: An Abstractive Model for Related Work Section Generation},
  author    = {Chen, Xiuying and Alamro, Hind and Li, Mingzhe and Gao, Shen and Zhang, Xiangliang and Zhao, Dongyan and Yan, Rui},
  booktitle = {Proceedings of the 59th Annual Meeting of the Association for Computational Linguistics and the 11th International Joint Conference on Natural Language Processing (Volume 1: Long Papers)},
  pages     = {6068--6077},
  year      = {2021}
}

@inproceedings{lu2020multi,
  title     = {{Multi-XScience}: A Large-Scale Dataset for Extreme Multi-Document Summarization of Scientific Articles},
  author    = {Lu, Yao and Dong, Yue and Charlin, Laurent},
  booktitle = {Proceedings of the 2020 Conference on Empirical Methods in Natural Language Processing (EMNLP)},
  pages     = {8068--8074},
  year      = {2020}
}

@inproceedings{kasanishi2023large,
  title     = {{SciReviewGen}: A Large-Scale Dataset for Automatic Literature Review Generation},
  author    = {Kasanishi, Tetsu and Isonuma, Masaru and Mori, Junichiro and Sakata, Ichiro},
  booktitle = {Findings of the Association for Computational Linguistics: ACL 2023},
  pages     = {6695--6715},
  year      = {2023},
  address   = {Toronto, Canada},
  publisher = {Association for Computational Linguistics}
}

@inproceedings{lewis2020retrieval,
  title     = {Retrieval-Augmented Generation for Knowledge-Intensive {NLP} Tasks},
  author    = {Lewis, Patrick and Perez, Ethan and Piktus, Aleksandra and Petroni, Fabio and Karpukhin, Vladimir and Goyal, Naman and K{\"u}ttler, Heinrich and Lewis, Mike and Yih, Wen-tau and Rockt{\"a}schel, Tim and Riedel, Sebastian and Kiela, Douwe},
  booktitle = {Advances in Neural Information Processing Systems},
  volume    = {33},
  pages     = {9459--9474},
  year      = {2020}
}

@article{gao2023retrieval,
  title   = {Retrieval-Augmented Generation for Large Language Models: A Survey},
  author  = {Gao, Yunfan and Xiong, Yun and Gao, Xinyu and Jia, Kangxiang and Pan, Jinliu and Bi, Yuxi and Dai, Yi and Sun, Jiawei and Wang, Haofen},
  journal = {arXiv preprint arXiv:2312.10997},
  year    = {2023}
}

@article{asai2026openscholar,
  title     = {Synthesizing Scientific Literature with Retrieval-Augmented Language Models},
  author    = {Asai, Akari and He, Jacqueline and Shao, Rulin and Shi, Weijia and Singh, Amanpreet and Chang, Joseph Chee and Lo, Kyle and Soldaini, Luca and Feldman, Sergey and D'Arcy, Mike and others},
  journal   = {Nature},
  pages     = {1--7},
  year      = {2026},
  publisher = {Nature Publishing Group}
}

@article{blei2003latent,
  title   = {Latent Dirichlet Allocation},
  author  = {Blei, David M. and Ng, Andrew Y. and Jordan, Michael I.},
  journal = {Journal of Machine Learning Research},
  volume  = {3},
  pages   = {993--1022},
  year    = {2003}
}

@inproceedings{mei2007automatic,
  title={Automatic Labeling of Multinomial Topic Models},
  author={Mei, Qiaozhu and Shen, Xuehua and Zhai, ChengXiang},
  booktitle={Proceedings of the 13th ACM SIGKDD International Conference on Knowledge Discovery and Data Mining},
  pages={490--499},
  year={2007}
}

@inproceedings{lau2011automatic,
  title={Automatic Labelling of Topic Models},
  author={Lau, Jey Han and Grieser, Karl and Newman, David and Baldwin, Timothy},
  booktitle={Proceedings of the 49th Annual Meeting of the Association for Computational Linguistics: Human Language Technologies},
  pages={1536--1545},
  year={2011}
}

@inproceedings{bhatia2016automatic,
  title     = {Automatic Labelling of Topics with Neural Embeddings},
  author    = {Bhatia, Shraey and Lau, Jey Han and Baldwin, Timothy},
  booktitle = {Proceedings of COLING 2016, the 26th International Conference on Computational Linguistics: Technical Papers},
  pages     = {953--963},
  year      = {2016}
}

@inproceedings{tang2025large,
  title={Large Language Models for Automated Literature Review: An Evaluation of Reference Generation, Abstract Writing, and Review Composition},
  author={Tang, Xuemei and Duan, Xufeng and Cai, Zhenguang G.},
  booktitle={Proceedings of the 2025 Conference on Empirical Methods in Natural Language Processing},
  pages={1602--1617},
  year={2025}
}

@inproceedings{liu2025llm,
  title={LLM-Guided Semantic-Aware Clustering for Topic Modeling},
  author={Liu, Jianghan and Shang, Ziyu and Ke, Wenjun and Wang, Peng and Luo, Zhizhao and Liu, Jiajun and Li, Guozheng and Li, Yining},
  booktitle={Proceedings of the 63rd Annual Meeting of the Association for Computational Linguistics},
  pages={18420--18435},
  year={2025},
  doi={10.18653/v1/2025.acl-long.902}
}

@inproceedings{yang2025neural,
  title     = {Neural Topic Modeling with Large Language Models in the Loop},
  author    = {Yang, Xiaohao and Zhao, He and Xu, Weijie and Qi, Yuanyuan and Lu, Jueqing and Phung, Dinh and Du, Lan},
  booktitle = {Proceedings of the 63rd Annual Meeting of the Association for Computational Linguistics (Volume 1: Long Papers)},
  pages     = {1377--1401},
  year      = {2025}
}

@article{blondel2008fast,
  title   = {Fast Unfolding of Communities in Large Networks},
  author  = {Blondel, Vincent D. and Guillaume, Jean-Loup and Lambiotte, Renaud and Lefebvre, Etienne},
  journal = {Journal of Statistical Mechanics: Theory and Experiment},
  volume  = {2008},
  number  = {10},
  pages   = {P10008},
  year    = {2008}
}

@article{nykvist2025testing,
  title={Testing the utility of GPT for title and abstract screening in environmental systematic evidence synthesis},
  author={Nykvist, Bj{\"o}rn and Macura, Biljana and Xylia, Maria and Olsson, Erik},
  journal={Environmental Evidence},
  volume={14},
  number={1},
  pages={7},
  year={2025},
  publisher={Springer}
}

@inproceedings{wang2025reassessing,
  title={Reassessing large language model boolean query generation for systematic reviews},
  author={Wang, Shuai and Scells, Harrisen and Koopman, Bevan and Zuccon, Guido},
  booktitle={Proceedings of the 48th International ACM SIGIR Conference on Research and Development in Information Retrieval},
  pages={3296--3305},
  year={2025}
}

@article{ye2024hybrid,
  title={A hybrid semi-automated workflow for systematic and literature review processes with large language model analysis},
  author={Ye, Anjia and Maiti, Ananda and Schmidt, Matthew and Pedersen, Scott J},
  journal={Future Internet},
  volume={16},
  number={5},
  pages={167},
  year={2024},
  publisher={MDPI}
}

@inproceedings{pei2025leveraging,
  title={Leveraging Llms for Streamlining and Demystifying the Systematic Literature Review Process},
  author={Pei, Bo and Sun, Xiaojiao},
  booktitle={2025 IEEE Frontiers in Education Conference (FIE)},
  pages={1--7},
  year={2025},
  organization={IEEE}
}

@inproceedings{silva2025enhancing,
  title={Enhancing systematic literature reviews: Evaluating the performance of LLM-based tools across key systematic literature review stages},
  author={Silva, Numaya and Wickramaarachchi, Dilani},
  booktitle={2025 5th International Conference on Advanced Research in Computing (ICARC)},
  pages={1--6},
  year={2025},
  organization={IEEE}
}

@article{padmakumar2025intent,
  title={Intent-aware Schema Generation And Refinement For Literature Review Tables},
  author={Padmakumar, Vishakh and Chang, Joseph Chee and Lo, Kyle and Downey, Doug and Naik, Aakanksha},
  journal={Findings of the Association for Computational Linguistics: EMNLP},
  volume={2025},
  pages={23450--23472},
  year={2025}
}

@article{hsu2024chime,
  title={Chime: Llm-assisted hierarchical organization of scientific studies for literature review support},
  author={Hsu, Chao-Chun and Bransom, Erin and Sparks, Jenna and Kuehl, Bailey and Tan, Chenhao and Wadden, David and Wang, Lucy Lu and Naik, Aakanksha},
  journal={Findings of the Association for Computational Linguistics: ACL 2024},
  pages={118--132},
  year={2024}
}

@inproceedings{john2026extractable,
  title={ExtracTable: Human-in-the-Loop Transformation of Scientific Corpora into Structured Knowledge},
  author={John, Lena and Ghanmi, Ahmed Malek and Wittenborg, Tim and Auer, S{\"o}ren and Karras, Oliver},
  booktitle={International Conference on Theory and Practice of Digital Libraries},
  pages={470--487},
  year={2026},
  organization={Springer}
}

@article{jansen2025data,
  title={Data extraction by generative artificial intelligence: Assessing determinants of accuracy using human-extracted data from systematic review databases},
  author={Jansen, Thorben and Liebenow, Lucas W and Mertens, Ute and Schmidt, Fabian TC and Lohmann, Julian F and Fleckenstein, Johanna and Meyer, Jennifer},
  journal={Psychological Bulletin},
  volume={151},
  number={10},
  pages={1280},
  year={2025},
  publisher={American Psychological Association}
}

@incollection{sarachuk2025still,
  title={Still not a remedy for academics: the use of generative {AI}-powered tools in bibliometric analysis},
  author={Sarachuk, Kirill},
  booktitle={Electronics, Communications and Networks},
  pages={57--63},
  year={2025},
  publisher={IOS Press}
}

@article{he2025remote,
  title={Remote sensing in river obstruction research: A bibliometric analysis integrated with large language model},
  author={He, Mingxia and Niu, Jie and Liu, Dongdong and Wu, Pan and Hu, Bill X},
  journal={Journal of Hydrology: Regional Studies},
  volume={62},
  pages={102850},
  year={2025},
  publisher={Elsevier}
}

@article{keenan2026beyond,
  title={Beyond manual review: using LLMs to systematically map five decades of {IFIP WG8.3} decision-support research},
  author={Keenan, Peter B and Heavin, Ciara},
  journal={Journal of Decision Systems},
  volume={35},
  number={1},
  pages={2665325},
  year={2026},
  publisher={Taylor \& Francis}
}

@article{yin2024quantitative,
  title={Quantitative analysis of scholars’ topic switching behavior in computer science: A two-dimensional metric approach},
  author={Yin, Dehu},
  journal={IEEE Access},
  volume={12},
  pages={104263--104271},
  year={2024},
  publisher={IEEE}
}

@article{donthu2021conduct,
  title={How to conduct a bibliometric analysis: An overview and guidelines},
  author={Donthu, Naveen and Kumar, Satish and Mukherjee, Debmalya and Pandey, Nitesh and Lim, Weng Marc},
  journal={Journal of Business Research},
  volume={133},
  pages={285--296},
  year={2021},
  publisher={Elsevier}
}

@article{zhang2019bertscore,
  title   = {BERTScore: Evaluating Text Generation with BERT},
  author  = {Zhang, Tianyi and Kishore, Varsha and Wu, Felix and Weinberger, Kilian Q. and Artzi, Yoav},
  journal = {arXiv preprint arXiv:1904.09675},
  year    = {2019}
}

\newpage

\appendix

\section{Ethical Considerations}

LLM-assisted bibliometric analysis may influence how researchers understand a field. Incorrect descriptions, unsupported references, or misleading cluster labels could distort scientific interpretation. We therefore evaluate reference grounding and distinguish in-corpus references from unsupported or invalid references.

The study uses bibliographic metadata and abstracts from Scopus. Licensing constraints may limit redistribution of raw records. Where raw Scopus data cannot be released, we will release prompts, code, derived identifiers, and evaluation scripts to support reproducibility within licensing constraints.

These results should be interpreted as evidence of structural fidelity, not as a general comparison between LLMs and human experts. Human descriptions were written for scholarly interpretation, not to optimize embedding-based coverage, ARI, or modularity. Thus, when structured LLM pipelines score higher on these metrics, it means they better preserve the reconstructed bibliometric or citation structure used in our evaluation, not that they offer greater domain insight, nuance, or scholarly value.

\section{Data collection and Pipeline Description}\label{app:pipelines}

\subsection{Data Collection}
\label{subsec:data-collection}
Before describing the pipelines, it is important to clarify that all pipelines required the number of clusters as an input parameter. This value was derived from the human-generated cluster descriptions, which served as the reference for comparison.

To construct this reference set, we collected 100 bibliometric studies, whose corpus characteristics are summarized in Table~\ref{tab:benchmark-summary}, and extracted the cluster descriptions reported in their bibliographic coupling analyses. For each PDF document, we generated a CSV file containing one row per cluster description. The total number of rows in each CSV file was then used to determine the number of clusters and was incorporated as an input parameter in the corresponding pipeline.

In addition, for each of the 100 studies, we extracted the search strategy used to construct the original corpus of papers. Specifically, we identified whether the study relied on Scopus or Web of Science, extracted the reported search query, and incorporated all available filters whenever possible. The query was subsequently executed in the corresponding database to retrieve the dataset of papers. When certain filters could not be implemented directly within the query, they were applied manually after retrieval.

Therefore, for each bibliometric study, three main elements were obtained: the human-generated cluster descriptions, the database query, and the corresponding dataset of papers. These elements were used as inputs across the pipelines, with the exception of Blind, which used only the query and the number of clusters derived from the human-generated cluster descriptions.

\begin{table}[t]
\centering
\caption{Summary of benchmark corpus characteristics.}
\label{tab:benchmark-summary}
\small
\begin{tabular}{p{0.30\linewidth}p{0.60\linewidth}}
\toprule
\textbf{Benchmark characteristic} & \textbf{Value} \\
\midrule
Top 5 domains &
\begin{itemize}[leftmargin=*, nosep]
    \item Innovation, entrepreneurship, and knowledge management
    \item AI, digital technology, and Industry 4.0
    \item Marketing, consumer behavior, and tourism
    \item Finance, accounting, and corporate governance
    \item Sustainability, environment, and climate
\end{itemize}
\\
\midrule
Scopus documents & Median: 492; IQR: 596 \\
Clusters & Median: 5; IQR: 2 \\
Publication year & Median: 2023.0; IQR: 3.25 \\
\bottomrule
\end{tabular}
\end{table}

\subsection{Pipeline Output}

Each pipeline produced a CSV file as output. This file contained the cluster identifier, the cluster description, and the references associated with each cluster.

\subsection{Pipelines}
\label{subsec:pipelines}
All six pipelines use the same base prompt, which defines the task, output structure, and formatting requirements. The complete prompt schema is reported in Appendix~\ref{app:prompt-schema}.

\subsubsection{Blind}

Blind used three inputs: the search query, the base prompt, and the target number of clusters. The query provided the thematic context of the study, while the base prompt specified the task instructions and the expected structure of the response. The target number of clusters defined the exact number of clusters to be generated. These inputs were passed to the LLM, which generated a structured response in JSON format. The JSON output was then parsed and saved as a CSV file containing the cluster identifier, cluster description, and associated references.

\subsubsection{Corpus}

Corpus used four inputs: the search query, the base prompt, the target number of clusters, and the Scopus/Web of Science corpus context. The query provided the thematic context of the study, while the base prompt specified the task instructions and the expected output format. The target number of clusters defined the exact number of clusters to be generated, and the corpus context provided the records used as the evidence base. These inputs were passed to the LLM, which generated a structured response in JSON format. The JSON output was then parsed and saved as a CSV file containing the cluster identifier, cluster description, and associated references.

\subsubsection{Corpus-select}

Corpus-select used four inputs: the search query, the base prompt, the target number of clusters, and the Scopus/Web of Science corpus context. The query provided the thematic context of the study, while the base prompt specified the task instructions. The target number of clusters defined the exact number of clusters to be generated, and the corpus context provided the records used as the evidence base.

This pipeline was implemented in two stages. First, the LLM identified the clusters and selected the most relevant references for each cluster, producing a structured list of cluster identifiers and associated paper identifiers. Second, the selected references were used to construct a reduced cluster-specific context. This focused context was then passed back to the LLM to generate the thematic description of each cluster. The final response was generated in JSON format, parsed, and saved as a CSV file containing the cluster identifier, cluster description, and associated references.

\subsection{Louvain Algorithm}

For Labeled, Labeled-select, and Ranked, the Scopus/Web of Science corpus context was enriched with cluster labels. These labels were generated using a bibliographic coupling and Louvain community detection procedure. In this approach, each paper was represented by its reference list, and papers were considered related when they cited the same prior works. The references were used to construct a paper-by-reference matrix, from which a bibliographic coupling matrix was computed. This matrix captured the degree of reference overlap between pairs of papers and was then normalized using association strength to obtain a similarity score.

The resulting similarities were used to build a weighted network, where nodes represented papers and edges represented bibliographic coupling relationships above a minimum threshold. Louvain community detection was then applied to identify clusters of papers. When human-generated cluster descriptions were available, the Louvain resolution parameter was adjusted so that the number of detected communities approximated the number of human clusters. After clustering, each paper received a cluster label, and a link strength value was computed to measure its centrality within the bibliographic coupling network. These labels were incorporated into the corpus context used as input for Labeled, Labeled-select, and Ranked.

\subsection{Label-Based Pipelines}

\subsubsection{Labeled}

Labeled used four inputs: the search query, the base prompt, the target number of clusters, and the labeled Scopus/Web of Science corpus context. The query provided the thematic context of the study, while the base prompt specified the task instructions and required output format. The target number of clusters indicated how many clusters had to be described, and the labeled corpus context provided the records together with their assigned cluster labels.

Using this labeled corpus context as the evidence base, the LLM generated one thematic description for each existing cluster while preserving the provided cluster identities. The final response was generated in JSON format, parsed, and saved as a CSV file containing the cluster identifier, cluster description, and associated references.

\subsubsection{Labeled-select}

Labeled-select used four inputs: the search query, the base prompt, the target number of clusters, and the labeled Scopus/Web of Science corpus context. The query provided the thematic context of the study, while the base prompt specified the task instructions. The target number of clusters indicated how many clusters had to be processed, and the labeled corpus context provided the records together with their assigned cluster labels.

This pipeline was implemented in two stages. First, the LLM selected the most relevant references within each labeled cluster, producing a structured list of cluster identifiers and associated paper identifiers. Second, the selected references were used to construct a reduced cluster-specific context. This focused context was then passed back to the LLM to generate one thematic description for each cluster while preserving the original cluster labels. The final response was generated in JSON format, parsed, and saved as a CSV file containing the cluster identifier, cluster description, and associated references.

\subsubsection{Ranked}

Ranked used four inputs: the search query, the base prompt, the target number of clusters, and the selected cluster context. The query provided the thematic context of the study, while the base prompt specified the task instructions and required output format. The target number of clusters indicated how many clusters had to be described, and the selected cluster context provided a reduced evidence set composed of the most representative papers from each labeled cluster.

Using this focused evidence set, the LLM generated one thematic description for each selected cluster while preserving the provided cluster labels. The final response was generated in JSON format, parsed, and saved as a CSV file containing the cluster identifier, cluster description, and associated references.

\section{Prompt Schema}
\label{app:prompt-schema}

\subsection{Blind}

\subsubsection{Prompt 1}

{\scriptsize\ttfamily
You are an expert research analyst in bibliometrics.\newline
\newline
This task is specifically based on bibliographic coupling, not general bibliometric analysis.\newline
\newline
Query:\newline
\{\{query\}\}\newline
\newline
Task:\newline
Identify exactly \{\{target\_cluster\_count\}\} coherent thematic clusters within this research topic.\newline
The number of clusters must exactly match the human reference clustering.\newline
Use cluster\_id values 1 through \{\{target\_cluster\_count\}\}.\newline
\newline
For each cluster:\newline
- write a precise academic description of the theme, main lines of inquiry, and distinctive focus\newline
- use less than 250 words to describe the cluster\newline
- provide references from your prior knowledge only\newline
- use no more than 10 references per cluster\newline
- do not browse or claim internet access\newline
- do not invent references if uncertain\newline
\newline
Output requirements:\newline
- return ONLY valid JSON\newline
- return a JSON array with exactly \{\{target\_cluster\_count\}\} objects\newline
- each object must contain: cluster\_id, description, references\newline
- references must be an array of bibliographic strings\newline
- cluster\_id must be numeric
}

\subsection{Corpus}

\subsubsection{Prompt 1}

{\scriptsize\ttfamily\raggedright
You are an expert in bibliometrics and scientific literature analysis.\newline
\newline
This task is specifically based on bibliographic coupling, not general bibliometric analysis.\newline
\newline
Query/context:\newline
\{\{query\}\}\newline
\newline
Use ONLY the Scopus records below.\newline
Identify exactly \{\{target\_cluster\_count\}\} thematic clusters supported by this corpus, describe each cluster, and select the most relevant papers for each cluster in the same response.\newline
The number of clusters must exactly match the human reference clustering.\newline
Use cluster\_id values 1 through \{\{target\_cluster\_count\}\}.\newline
Use less than 250 words to describe each cluster.\newline
\newline
Reference rules:\newline
- every reference must be written exactly as [\#]\newline
- \# must correspond to one of the provided paper\_id values\newline
- only cite papers from the provided Scopus records\newline
- only select papers that are truly relevant for describing the cluster\newline
- use no more than 10 references per cluster\newline
\newline
Output requirements:\newline
- return ONLY valid JSON\newline
- return a JSON array with exactly \{\{target\_cluster\_count\}\} objects\newline
- each object must contain: cluster\_id, description, references\newline
- references must be an array of [\#] tokens\newline
\newline
Scopus records:\newline
\{\{scopus\_context\}\}
\par}

\subsection{Corpus-select}

\subsubsection{Prompt 1}

{\scriptsize\ttfamily\raggedright
You are an expert in bibliometrics and scientific literature analysis.\newline
\newline
This task is specifically based on bibliographic coupling, not general bibliometric analysis.\newline
\newline
Query/context:\newline
\{\{query\}\}\newline
\newline
Use ONLY the Scopus records below.\newline
Identify exactly \{\{target\_cluster\_count\}\} thematic clusters and select the most relevant papers for each cluster.\newline
The number of clusters must exactly match the human reference clustering.\newline
Use cluster\_id values 1 through \{\{target\_cluster\_count\}\}.\newline
\newline
Reference rules:\newline
- every reference must be written exactly as [\#]\newline
- \# must correspond to one of the provided paper\_id values\newline
- only select papers that are truly relevant for describing the cluster\newline
- use no more than 10 references per cluster\newline
\newline
Output requirements:\newline
- return ONLY valid JSON\newline
- return a JSON array with exactly \{\{target\_cluster\_count\}\} objects\newline
- each object must contain: cluster\_id, references\newline
- references must be an array of [\#] tokens\newline
\newline
Scopus records:\newline
\{\{scopus\_context\}\}
\par}

\subsubsection{Prompt 2}

{\scriptsize\ttfamily\raggedright
You are an expert in bibliometrics and scientific literature analysis.\newline
\newline
This task is specifically based on bibliographic coupling, not general bibliometric analysis.\newline
\newline
Query/context:\newline
\{\{query\}\}\newline
\newline
Use ONLY the selected cluster papers below.\newline
Write one academic description for each of the exactly \{\{target\_cluster\_count\}\} clusters.\newline
Use cluster\_id values 1 through \{\{target\_cluster\_count\}\}.\newline
Use less than 250 words to describe each cluster.\newline
\newline
Output requirements:\newline
- return ONLY valid JSON\newline
- return a JSON array with exactly \{\{target\_cluster\_count\}\} objects\newline
- each object must contain: cluster\_id, description\newline
- do not include references in this step\newline
\newline
Selected cluster papers:\newline
\{\{cluster\_contexts\}\}
\par}

\subsection{Labeled}

\subsubsection{Prompt 1}

{\scriptsize\ttfamily\raggedright
You are an expert in bibliometrics and scientific literature analysis.\newline
\newline
This task is specifically based on bibliographic coupling, not general bibliometric analysis.\newline
\newline
Query/context:\newline
\{\{query\}\}\newline
\newline
Use ONLY the labeled Scopus records below.\newline
Each paper already has a cluster label, and you must preserve those labels.\newline
Write one academic description for each of the exactly \{\{target\_cluster\_count\}\} labeled clusters using the whole labeled corpus as context.\newline
Use less than 250 words to describe each cluster.\newline
\newline
Reference rules:\newline
- every reference must be written exactly as [\#]\newline
- use only the provided paper\_id values\newline
- do not cite papers outside the labeled Scopus records\newline
- use no more than 10 references per cluster\newline
\newline
Output requirements:\newline
- return ONLY valid JSON\newline
- return a JSON array with exactly \{\{target\_cluster\_count\}\} objects\newline
- each object must contain: cluster\_id, description, references\newline
- cluster\_id must match one of the provided labeled cluster values\newline
- references must be an array of [\#] tokens\newline
\newline
Labeled Scopus records:\newline
\{\{labeled\_scopus\_context\}\}
\par}

\subsection{Labeled-select}

\subsubsection{Prompt 1}

{\scriptsize\ttfamily\raggedright
You are an expert in bibliometrics and scientific literature analysis.\newline
\newline
This task is specifically based on bibliographic coupling, not general bibliometric analysis.\newline
\newline
Query/context:\newline
\{\{query\}\}\newline
\newline
Use ONLY the labeled Scopus records below.\newline
Each paper already has a cluster label, and you must preserve those labels.\newline
Select the papers that are most relevant for writing a representative description for each of the exactly \{\{target\_cluster\_count\}\} labeled clusters.\newline
\newline
Reference rules:\newline
- every reference must be written exactly as [\#]\newline
- use only the provided paper\_id values\newline
- for each cluster\_id, only select papers that belong to that same labeled cluster\newline
- use no more than 10 references per cluster\newline
\newline
Output requirements:\newline
- return ONLY valid JSON\newline
- return a JSON array with exactly \{\{target\_cluster\_count\}\} objects\newline
- each object must contain: cluster\_id, references\newline
- cluster\_id must match one of the provided labeled cluster values\newline
- references must be an array of [\#] tokens\newline
\newline
Labeled Scopus records:\newline
\{\{labeled\_scopus\_context\}\}
\par}

\subsubsection{Prompt 2}

{\scriptsize\ttfamily\raggedright
You are an expert in bibliometrics and scientific literature analysis.\newline
\newline
This task is specifically based on bibliographic coupling, not general bibliometric analysis.\newline
\newline
Query/context:\newline
\{\{query\}\}\newline
\newline
Use ONLY the labeled Scopus records below.\newline
Each paper already has a cluster label, and you must preserve those labels.\newline
Select the papers that are most relevant for writing a representative description for each of the exactly \{\{target\_cluster\_count\}\} labeled clusters.\newline
\newline
Reference rules:\newline
- every reference must be written exactly as [\#]\newline
- use only the provided paper\_id values\newline
- for each cluster\_id, only select papers that belong to that same labeled cluster\newline
- use no more than 10 references per cluster\newline
\newline
Output requirements:\newline
- return ONLY valid JSON\newline
- return a JSON array with exactly \{\{target\_cluster\_count\}\} objects\newline
- each object must contain: cluster\_id, references\newline
- cluster\_id must match one of the provided labeled cluster values\newline
- references must be an array of [\#] tokens\newline
\newline
Labeled Scopus records:\newline
\{\{labeled\_scopus\_context\}\}
\par}

\subsection{Ranked}

\subsubsection{Prompt 1}

{\scriptsize\ttfamily\raggedright
You are an expert research analyst writing bibliometric cluster descriptions.\newline
\newline
This task is specifically based on bibliographic coupling, not general bibliometric analysis.\newline
\newline
Query/context:\newline
\{\{query\}\}\newline
\newline
You will be provided multiple bibliographic coupling clusters at once.\newline
For each cluster, you will receive the top papers of that cluster ranked by link strength.\newline
Use ONLY the selected cluster papers below, and preserve the provided cluster labels.\newline
\newline
Your task:\newline
Write one coherent academic description for each of the exactly \{\{target\_cluster\_count\}\} clusters.\newline
\newline
Strict rules:\newline
- use ONLY the information provided below\newline
- do NOT introduce external knowledge or invent papers, methods, findings, datasets, journals, years, or topics\newline
- base all statements strictly on patterns visible in the provided titles and abstracts\newline
- identify the main research theme of each cluster\newline
- emphasize the topic, subtheme, angle, or intellectual profile that makes each cluster distinctive relative to the others in this set\newline
- do not infer distinctions that are not clearly supported by the provided information\newline
- use less than 250 words to describe each cluster\newline
- every reference must be written exactly as [\#]\newline
- use only the provided paper\_id values\newline
- each cluster description must cite at least 3 different papers from that same cluster when enough papers are available\newline
- use no more than 10 references per cluster\newline
- write each cluster description as one dense academic paragraph or two short paragraphs\newline
\newline
Output requirements:\newline
- return ONLY valid JSON\newline
- return a JSON array with exactly \{\{target\_cluster\_count\}\} objects\newline
- each object must contain: cluster\_id, description, references\newline
- cluster\_id must match one of the provided cluster labels\newline
- references must be an array of [\#] tokens\newline
\newline
Selected cluster papers:\newline
\{\{selected\_clusters\_context\}\}
\par}

\begin{figure*}[t]
    \centering
    \includegraphics[width=1\linewidth]{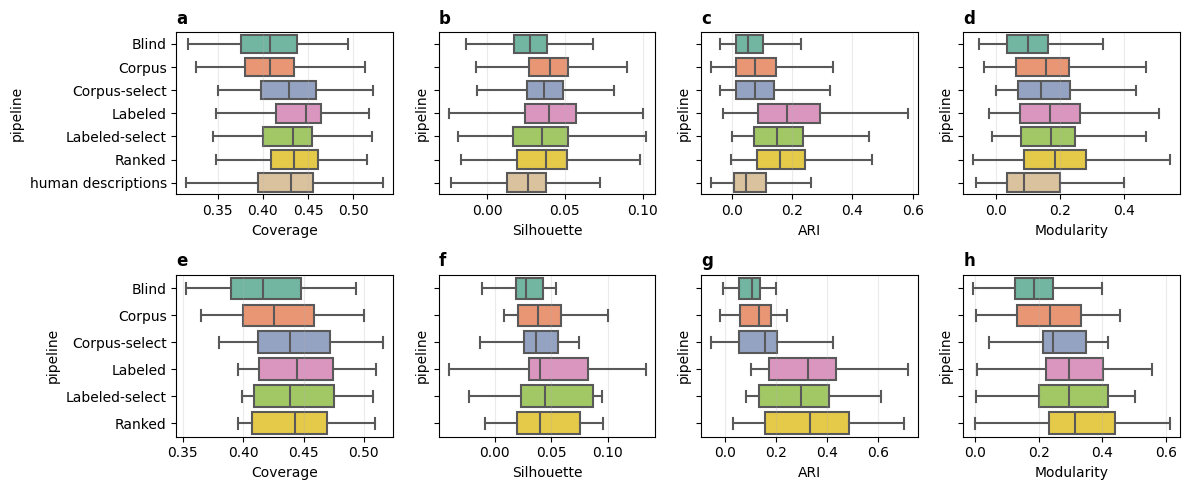}
    \caption{Boxplots of pipeline performance across papers under the bibliographic and citation settings. Panels (a–d) correspond to the bibliographic setting and panels (e–h) to the citation setting}
    \label{fig:placeholder}
\end{figure*}

\section{Evaluation}\label{app:evaluation-details}

\subsection{Human Comparison}
We compare generated cluster descriptions with the human-authored benchmark descriptions using BERTScore \citep{zhang2019bertscore}. For each source paper \(r_i\), the human benchmark is

\[
H_i = \{h_{i1}, h_{i2}, \ldots, h_{iK_i}\},
\]

and the pipeline output is

\[
Y_i =
\{(l_{i1}, d_{i1}), \ldots, (l_{iK_i}, d_{iK_i})\},
\]

where \(d_{ik}\) is the generated description for cluster \(k\). We compute the BERTScore F1 value between every generated description \(d_{iu}\) and every human description \(h_{iv}\), producing an alignment matrix \(A_i \in \mathbb{R}^{K_i \times K_i}\):

\[
A_{iuv}
=
\operatorname{BERTScore}_{F1}(d_{iu}, h_{iv}).
\]

We then solve a one-to-one assignment problem that maximizes the total semantic alignment between generated and human descriptions. Let \(M_i^*\) denote the optimal set of matched generated--human pairs. Since each generated output contains \(K_i\) clusters, the human-comparison score for source paper \(r_i\) is the mean BERTScore F1 across the \(K_i\) optimally matched pairs:

\[
\operatorname{HumanBERT}(i)
=
\frac{1}{K_i}
\sum_{(u,v) \in M_i^*}
A_{iuv}.
\]

Higher values indicate that the generated cluster descriptions are more semantically similar to the human-authored benchmark descriptions.

\subsection{Semantic Evaluation}

The semantic evaluation asks whether the generated cluster descriptions collectively cover the content of the reconstructed Scopus corpus \(D_i\). Let

\[
S_i = \{s_{i1}, \ldots, s_{ia}\}
\]

be the set of sentence-level textual atoms extracted from \(D_i\), and let

\[
B_i = \{b_{i1}, \ldots, b_{ir}\}
\]

be the set of textual atoms extracted from the generated cluster descriptions.

We embed all corpus and description atoms into a shared embedding space. For each corpus sentence \(s_{ij}\), we compute its maximum cosine similarity to any generated description atom:

\[
\operatorname{cov}(s_{ij}) =
\max_{b \in B_i}
\cos(e(s_{ij}), e(b)).
\]

The semantic coverage score for instance \(i\) is:

\[
\operatorname{Coverage}(i) =
\frac{1}{|S_i|}
\sum_{s_{ij} \in S_i}
\operatorname{cov}(s_{ij}).
\]

This metric measures how well the generated descriptions collectively cover the semantic content of the corpus. We treat it as a scalable proxy for coverage rather than as a direct measure of factual entailment.

\subsection{Clustering Evaluation}

The clustering evaluation asks whether the generated descriptions induce paper-level assignments that recover the bibliometric or citation clustering \(Z_i\). For each generated cluster \(k\), let \(d_{ik}\) be the text representation formed by concatenating its label and description. For each paper \(p_{ij}\), let \(a_{ij}\) be its title-and-abstract representation. We assign each paper to the most similar generated cluster:

\[
\hat{z}_{ij}
=
\arg\max_{k \in \{1,\ldots,K_i\}}
\cos(e(a_{ij}), e(d_{ik})).
\]

This produces an induced partition \(\hat{Z}_i\) of the reconstructed corpus. Our primary clustering metric is Adjusted Rand Index (ARI) \citep{hubert1985comparing}, computed between the induced partition \(\hat{Z}_i\) and the Louvain bibliometric or citation clustering \(Z_i\):

\[
\operatorname{ARI}(\hat{Z}_i, Z_i).
\]

ARI measures whether the paper groups implied by the generated descriptions match the clusters produced by the standard bibliometric clustering procedure. For corpus-based pipelines, this evaluates whether the LLM recovers the algorithmic cluster structure from text alone. For labeled and ranked pipelines, it evaluates whether the generated descriptions preserve the meaning of the clusters they were given.

As a secondary clustering metric, we compute the silhouette score \citep{rousseeuw1987silhouettes} of the induced partition \(\hat{Z}_i\) in abstract embedding space. The silhouette score measures whether papers assigned to the same generated cluster are semantically closer to one another than to papers assigned to different clusters. While ARI measures agreement with the bibliometric clustering, silhouette measures semantic separability of the induced groups.

\subsection{Graph Evaluation}

The graph evaluation asks whether the induced partition \(\hat{Z}_i\) respects the bibliographic or citation structure of the corpus. This is distinct from ARI: ARI compares the induced assignments to the Louvain clustering \(Z_i\), while graph modularity directly measures how well the induced assignments align with the underlying paper-relation structure.

Given induced assignments \(\hat{Z}_i\), we compute modularity \citep{newman2004modularity}:

\[
Q =
\frac{1}{2w}
\sum_{u,v}
\left[
w_{uv} -
\frac{k_u k_v}{2w}
\right]
\mathbb{I}[\hat{z}_u = \hat{z}_v],
\]

where \(w_{uv}\) is the edge weight between papers \(u\) and \(v\), \(k_u\) is the weighted degree of paper \(u\), and \(2w = \sum_{u,v} w_{uv}\).

Higher modularity indicates that the clusters induced by the generated descriptions correspond more strongly to dense regions of the bibliographic or citation relation structure.

\subsection{Reference Grounding}

We evaluate reference grounding in two ways. For \textit{Blind}, we validate whether generated references correspond to real bibliographic records. Because this pipeline may generate references without access to the reconstructed Scopus corpus, each reference is matched to the closest OpenAlex record and evaluated under three increasingly strict criteria: title match, title plus publication year, and title plus publication year plus first author. Title matching uses fuzzy string similarity with a threshold of 80, year matching requires the same publication year, and author matching requires the same normalized first-author surname.

For the other pipelines, we evaluate whether generated references are grounded in the provided evidence. Each generated reference is classified as in-corpus valid, out-of-corpus valid, or invalid. In-corpus valid references point to papers in the reconstructed Scopus corpus \(D_i\). Out-of-corpus valid references correspond to real papers outside \(D_i\), while invalid references cannot be matched to a real paper or valid corpus identifier. This distinction is especially important because less structured pipelines may rely on parametric memory, whereas structured pipelines constrain references to retrieved papers, algorithmic clusters, or link-ranked evidence.

We also measure whether each generated description reflects the cited evidence. Reference-grounded coverage applies the same sentence-level coverage procedure used in the semantic evaluation, but restricts the evidence set to the papers explicitly cited by each generated cluster. For each source paper \(r_i\) and generated cluster \(k\), let \(E_{ik}^{\mathrm{ref}}\) be the set of sentence-level textual atoms extracted from the Scopus abstracts cited in that cluster's references field, and let \(B_{ik}\) be the set of sentence-level textual atoms extracted from the generated description \(d_{ik}\). The cluster-level score is

\[
\operatorname{RGC}_{ik}
=
\frac{1}{|E_{ik}^{\mathrm{ref}}|}
\sum_{e \in E_{ik}^{\mathrm{ref}}}
\max_{b \in B_{ik}}
\cos(\mathbf{e}(e), \mathbf{e}(b)).
\]

The final score for source paper \(r_i\) is the average across generated clusters:

\[
\operatorname{RGC}(i)
=
\frac{1}{K_i}
\sum_{k=1}^{K_i}
\operatorname{RGC}_{ik}.
\]

Higher values indicate that generated cluster descriptions more fully reflect the semantic content of the specific Scopus abstracts they cite as supporting evidence.

\subsection{Evaluation rationale}
Our evaluation is designed around the purpose of bibliometric analysis, not generic text-generation quality. A bibliometric cluster description should interpret a science map by covering the corpus, distinguishing the induced clusters, aligning with the underlying paper-relation graph, and citing supporting evidence. We therefore complement semantic similarity to human-written descriptions with structure-sensitive metrics: semantic coverage, ARI, silhouette, modularity, and reference grounding.

Human-authored descriptions remain an important reference for interpretive alignment, but they are not the sole evaluation target. Published cluster descriptions are written for scholarly explanation and may emphasize conceptual nuance or narrative framing rather than paper-level reconstruction in a rebuilt corpus. Accordingly, we use human comparison as one dimension of evaluation, while the remaining metrics test whether generated descriptions preserve the bibliometric structure they are intended to explain.

While our evaluation captures several important aspects of bibliometric cluster description quality, it also highlights limitations that motivate the need for a broader evaluation rationale. In particular, this work should be positioned not only as a bibliometric study, but also as an NLP evaluation of structured LLM-assisted scientific synthesis. The proposed metrics assess semantic alignment, corpus coverage, clustering fidelity, graph consistency, and reference grounding, but they do not fully capture expert judgment, conceptual nuance, or the scholarly usefulness of a generated cluster description. In addition, the benchmark reconstruction process and human-evaluation protocol require careful reporting, since differences between reconstructed corpora and the original bibliometric studies may affect downstream comparisons. These considerations motivate our use of multiple complementary metrics and our interpretation of the results as evidence of structural fidelity rather than as a direct replacement for human bibliometric expertise.

\section{Ablations}
\label{sec:ablations}

To further analyze the behavior of the proposed pipelines, a set of ablation studies was conducted. These experiments were not performed on the full set of 100 bibliometric studies, but on a subset of 20 studies, due to the computational and financial cost associated with repeatedly executing the pipelines under different configurations.

\subsection{Generative Model Ablation}
The first ablation study evaluated the effect of the language model used to generate the cluster descriptions. In this experiment, the pipelines were executed using two different generative models: GPT-5.4 and Gemini-2.5-Flash. The objective was to compare how the performance and behavior of the pipelines changed depending on the model responsible for producing the thematic cluster descriptions.

Table~\ref{tab:model-source-results} reports the results of the generative-model ablation. The results show broadly consistent behavior across the two generative models. The \textit{Blind} pipeline remains the weakest setting for most metrics, while the best results are generally obtained by the structured pipelines, especially \textit{Labeled} and \textit{Ranked}. Under GPT-5.4, \textit{Labeled} achieves the best mean-rank and median coverage results, as well as the strongest median ARI and modularity. Under Gemini-2.5-Flash, \textit{Ranked} obtains the best median modularity and nearly the best median ARI, while \textit{Labeled} has the best ARI mean rank and highest ARI win percentage. The main difference between models is therefore not whether structure helps, but which structured pipeline benefits most. This supports the robustness of the main finding that LLM-assisted bibliometric synthesis performs best when the model is given algorithmic cluster structure or representative cluster evidence.

\begin{table*}[t]
\centering
\caption{Benchmark results by model source. Results are grouped by summary statistic. Median summarizes central performance, mean rank summarizes relative ordering across instances, and win percentage reports how often each pipeline achieves the best score. Bold values indicate the best result for each metric and model source within each statistic.}
\label{tab:model-source-results}
\small
\resizebox{\textwidth}{!}{%
\begin{tabular}{lcccccccc}
\toprule
\textbf{Pipeline}
& \multicolumn{2}{c}{\textbf{Semantic}}
& \multicolumn{4}{c}{\textbf{Cluster quality}}
& \multicolumn{2}{c}{\textbf{Graph quality}} \\
\cmidrule(lr){2-3}
\cmidrule(lr){4-7}
\cmidrule(lr){8-9}
& \multicolumn{2}{c}{\textbf{Coverage}}
& \multicolumn{2}{c}{\textbf{Silhouette}}
& \multicolumn{2}{c}{\textbf{ARI}}
& \multicolumn{2}{c}{\textbf{Modularity}} \\
\cmidrule(lr){2-3}
\cmidrule(lr){4-5}
\cmidrule(lr){6-7}
\cmidrule(lr){8-9}
& \textbf{GPT-5.4} & \textbf{Gemini-2.5-Flash}
& \textbf{GPT-5.4} & \textbf{Gemini-2.5-Flash}
& \textbf{GPT-5.4} & \textbf{Gemini-2.5-Flash}
& \textbf{GPT-5.4} & \textbf{Gemini-2.5-Flash} \\
\midrule

\multicolumn{9}{l}{\textit{Mean rank}} \\
\midrule
Blind
& 5.85 & 6.60
& 5.10 & 5.30
& 5.55 & 5.85
& 5.40 & 5.45 \\

Corpus
& 6.30 & 5.20
& 4.10 & \textbf{3.10}
& 4.90 & 4.65
& 3.70 & 3.90 \\

Corpus-select
& 3.55 & 3.70
& 3.35 & 4.30
& 4.55 & 4.85
& \textbf{3.00} & 4.30 \\

Labeled
& \textbf{1.90} & 3.05
& 3.35 & 3.50
& \textbf{2.25} & \textbf{1.95}
& 3.90 & 3.30 \\

Labeled-select
& 3.95 & 3.55
& 4.10 & 3.75
& 3.15 & 2.55
& 4.10 & 3.50 \\

Ranked
& 2.65 & \textbf{2.05}
& \textbf{3.30} & 3.60
& 2.35 & 2.75
& 3.10 & \textbf{2.80} \\

Human
& 3.80 & 3.85
& 4.70 & 4.45
& 5.25 & 5.40
& 4.80 & 4.75 \\

\midrule
\multicolumn{9}{l}{\textit{Median}} \\
\midrule
Blind
& 0.4085 & 0.4054
& 0.0276 & 0.0211
& 0.0496 & 0.0463
& 0.1082 & 0.0660 \\

Corpus
& 0.4021 & 0.4181
& 0.0395 & 0.0437
& 0.0700 & 0.0556
& 0.1617 & 0.1582 \\

Corpus-select
& 0.4234 & 0.4299
& 0.0369 & 0.0322
& 0.0650 & 0.0630
& 0.1422 & 0.1428 \\

Labeled
& \textbf{0.4512} & 0.4324
& \textbf{0.0436} & \textbf{0.0494}
& \textbf{0.1661} & 0.1762
& \textbf{0.2071} & 0.1983 \\

Labeled-select
& 0.4201 & 0.4230
& 0.0402 & 0.0360
& 0.1190 & 0.1483
& 0.1928 & 0.1898 \\

Ranked
& 0.4280 & 0.4321
& 0.0421 & 0.0453
& 0.1299 & \textbf{0.1763}
& 0.2059 & \textbf{0.2103} \\

Human
& 0.4336 & \textbf{0.4336}
& 0.0319 & 0.0319
& 0.0438 & 0.0438
& 0.1113 & 0.1113 \\

\midrule
\multicolumn{9}{l}{\textit{Win percentage}} \\
\midrule
Blind
& 0.00 & 0.00
& 0.00 & 0.00
& 5.00 & 0.00
& 0.00 & 10.00 \\

Corpus
& 0.00 & 0.00
& \textbf{20.00} & 20.00
& 0.00 & 5.00
& 15.00 & 5.00 \\

Corpus-select
& 5.00 & 15.00
& 15.00 & 10.00
& 10.00 & 0.00
& \textbf{30.00} & 5.00 \\

Labeled
& \textbf{35.00} & 0.00
& 10.00 & \textbf{25.00}
& \textbf{35.00} & \textbf{50.00}
& 10.00 & \textbf{25.00} \\

Labeled-select
& 10.00 & 0.00
& \textbf{20.00} & 15.00
& 20.00 & 30.00
& 15.00 & 20.00 \\

Ranked
& 25.00 & \textbf{50.00}
& \textbf{20.00} & 10.00
& 30.00 & 15.00
& 25.00 & \textbf{25.00} \\

Human
& 25.00 & 35.00
& 15.00 & 20.00
& 0.00 & 0.00
& 5.00 & 10.00 \\

\bottomrule
\end{tabular}%
}
\end{table*}

\begin{figure*}[t]
    \centering

    \begin{minipage}[t]{0.22\textwidth}
        \centering
        \includegraphics[width=\linewidth]{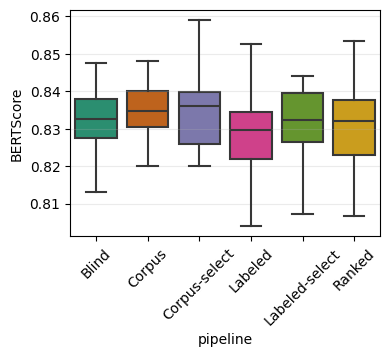}
        \caption{BERTScore human alignment by pipeline for Gemini.}
        \label{fig:BERTScore-gemini}
    \end{minipage}
    \hfill
    \begin{minipage}[t]{0.22\textwidth}
        \centering
        \includegraphics[width=\linewidth]{reference-grounded_coverage.png}
        \caption{Distribution of reference-grounded coverage scores across description-generation pipelines using Gemini.}
        \label{fig:reference_grounded_coverage-gemini}
    \end{minipage}
    \hfill
    \begin{minipage}[t]{0.54\textwidth}
        \centering
        \includegraphics[width=\linewidth]{reference_grounding.png}
        \caption{Histograms of cluster-level reference precision under progressively stricter matching criteria for Gemini. Panel (a) shows precision based on title matching only, panel (b) requires both title and year to match, and panel (c) requires title, year, and first author to match.}
        \label{fig:cluster-level-ref-gemini}
    \end{minipage}

\end{figure*}

\subsection{Embedding Model Ablation}

The second ablation study tested whether the evaluation results depended on the embedding model used to compute semantic similarities. We evaluated the same generated cluster descriptions using two embedding models: \texttt{all-mpnet-base-v2} and \texttt{text-embedding-3-large}. All other experimental settings were kept fixed. We then recomputed the evaluation metrics reported in Table~\ref{tab:embedding-model-results}: coverage, silhouette score, ARI, and modularity. Stable relative performance across both embedding models was interpreted as evidence that the pipeline comparisons were robust to the choice of semantic representation model.

Table~\ref{tab:embedding-model-results} shows that the main conclusions are broadly stable across embedding models. In both MPNet and OpenAI embedding spaces, the \textit{Blind} pipeline remains among the weakest settings, while the structured pipelines, especially \textit{Labeled} and \textit{Ranked}, obtain the strongest results across ARI and modularity. \textit{Labeled} is the most consistently strong pipeline, achieving the best mean-rank and median results for coverage and ARI under both embedding models, as well as the best median modularity. \textit{Ranked} is also competitive, especially for silhouette and modularity. The main difference is that MPNet gives higher absolute coverage and silhouette values than OpenAI, but the relative ordering of the pipelines remains similar. This suggests that the evaluation results are not driven by a single embedding model.

\begin{table*}[t]
\centering
\caption{Benchmark results by embedding model. MPNet denotes \texttt{all-mpnet-base-v2}, and OpenAI denotes \texttt{text-embedding-3-large}. Median summarizes central performance, mean rank summarizes relative ordering across instances, and win percentage reports how often each pipeline achieves the best score. Bold values indicate the best result for each metric and embedding model within each statistic.}
\label{tab:embedding-model-results}
\small
\resizebox{\textwidth}{!}{%
\begin{tabular}{lcccccccc}
\toprule
\textbf{Pipeline}
& \multicolumn{2}{c}{\textbf{Semantic}}
& \multicolumn{4}{c}{\textbf{Cluster quality}}
& \multicolumn{2}{c}{\textbf{Graph quality}} \\
\cmidrule(lr){2-3}
\cmidrule(lr){4-7}
\cmidrule(lr){8-9}
& \multicolumn{2}{c}{\textbf{Coverage}}
& \multicolumn{2}{c}{\textbf{Silhouette}}
& \multicolumn{2}{c}{\textbf{ARI}}
& \multicolumn{2}{c}{\textbf{Modularity}} \\
\cmidrule(lr){2-3}
\cmidrule(lr){4-5}
\cmidrule(lr){6-7}
\cmidrule(lr){8-9}
& \textbf{MPNet} & \textbf{OpenAI}
& \textbf{MPNet} & \textbf{OpenAI}
& \textbf{MPNet} & \textbf{OpenAI}
& \textbf{MPNet} & \textbf{OpenAI} \\
\midrule

\multicolumn{9}{l}{\textit{Mean rank}} \\
\midrule
Blind
& 5.40 & 5.85
& 5.05 & 5.10
& 5.65 & 5.55
& 5.00 & 5.40 \\

Corpus
& 6.20 & 6.30
& 3.70 & 4.10
& 4.60 & 4.90
& 4.40 & 3.70 \\

Corpus-select
& 3.70 & 3.55
& 3.70 & 3.35
& 5.25 & 4.55
& 3.75 & \textbf{3.00} \\

Labeled
& \textbf{1.75} & \textbf{1.90}
& \textbf{3.50} & 3.35
& \textbf{1.90} & \textbf{2.25}
& \textbf{2.90} & 3.90 \\

Labeled-select
& 4.35 & 3.95
& 4.20 & 4.10
& 3.00 & 3.15
& 3.75 & 4.10 \\

Ranked
& 3.35 & 2.65
& 3.65 & \textbf{3.30}
& 2.10 & 2.35
& 3.10 & 3.10 \\

Human
& 3.25 & 3.80
& 4.20 & 4.70
& 5.50 & 5.25
& 5.10 & 4.80 \\

\midrule
\multicolumn{9}{l}{\textit{Median}} \\
\midrule
Blind
& 0.4585 & 0.4085
& 0.0393 & 0.0276
& 0.0547 & 0.0496
& 0.1190 & 0.1082 \\

Corpus
& 0.4524 & 0.4021
& 0.0604 & 0.0395
& 0.0600 & 0.0700
& 0.1319 & 0.1617 \\

Corpus-select
& 0.4812 & 0.4234
& 0.0559 & 0.0369
& 0.0580 & 0.0650
& 0.1278 & 0.1422 \\

Labeled
& \textbf{0.4988} & \textbf{0.4512}
& 0.0633 & \textbf{0.0436}
& \textbf{0.1863} & \textbf{0.1661}
& \textbf{0.2063} & \textbf{0.2071} \\

Labeled-select
& 0.4670 & 0.4201
& 0.0607 & 0.0402
& 0.1291 & 0.1190
& 0.1903 & 0.1928 \\

Ranked
& 0.4688 & 0.4280
& \textbf{0.0687} & 0.0421
& 0.1461 & 0.1299
& 0.1913 & 0.2059 \\

Human
& 0.4834 & 0.4336
& 0.0563 & 0.0319
& 0.0552 & 0.0438
& 0.1053 & 0.1113 \\

\midrule
\multicolumn{9}{l}{\textit{Win percentage}} \\
\midrule
Blind
& 0.00 & 0.00
& 5.00 & 0.00
& 0.00 & 5.00
& 5.00 & 0.00 \\

Corpus
& 0.00 & 0.00
& 15.00 & \textbf{20.00}
& 5.00 & 0.00
& 0.00 & 15.00 \\

Corpus-select
& 10.00 & 5.00
& 20.00 & 15.00
& 0.00 & 10.00
& 10.00 & \textbf{30.00} \\

Labeled
& 40.00 & \textbf{35.00}
& \textbf{35.00} & 10.00
& \textbf{60.00} & \textbf{35.00}
& 30.00 & 10.00 \\

Labeled-select
& 0.00 & 10.00
& 5.00 & \textbf{20.00}
& 15.00 & 20.00
& 10.00 & 15.00 \\

Ranked
& 0.00 & 25.00
& 10.00 & \textbf{20.00}
& 20.00 & 30.00
& \textbf{35.00} & 25.00 \\

Human
& \textbf{50.00} & 25.00
& 10.00 & 15.00
& 0.00 & 0.00
& 10.00 & 5.00 \\

\bottomrule
\end{tabular}%
}
\end{table*}

\subsection{Prompt Ablation}

The third ablation study evaluated the effect of prompt formulation on cluster-description generation. We compared a simpler prompt against an enriched prompt that included two additional elements. First, the enriched prompt added the query field:

\begin{quote}
\ttfamily
Query/context: \{\{query\}\}
\end{quote}

Second, it added a methodological explanation of the bibliographic relation:

\begin{quote}
\ttfamily
The analysis that will be performed is a bibliographic coupling analysis. Bibliographic coupling clusters papers that share cited references. Papers in the same cluster often reflect related intellectual backgrounds, methods, or problem framings.
\end{quote}

The simpler prompt removed both of these elements. This ablation tests whether giving the model explicit topic context and bibliometric-relation context improves the generated descriptions relative to a more minimal prompt.

Table~\ref{tab:prompt-configuration-results} shows that prompt formulation affects performance, but it does not change the main pattern across pipelines. The \textit{Blind} pipeline remains weak across most configurations, while structured pipelines continue to perform best. Adding bibliographic-coupling context improves some structural results, especially for \textit{Labeled}, which achieves the best ARI mean rank and win percentage under the bibliographic prompt, and for \textit{Ranked}, which obtains the best modularity median under the same configuration. The no-query condition improves coverage for \textit{Corpus-select} and modularity for \textit{Labeled-select}, but these gains are not consistent across all metrics. Overall, the results suggest that prompt details can shift which structured pipeline performs best, but they do not overturn the broader conclusion that externally provided bibliometric structure is more important than prompt formulation alone.

\begin{table*}[t]
\centering
\caption{Benchmark results by prompt configuration. \textit{Biblio.} denotes prompts with bibliographic context. Median summarizes central performance, mean rank summarizes relative ordering across instances, and win percentage reports how often each pipeline achieves the best score. Bold values indicate the best result for each metric and prompt configuration within each statistic.}
\label{tab:prompt-configuration-results}
\small
\resizebox{\textwidth}{!}{%
\begin{tabular}{lcccccccccccc}
\toprule
\textbf{Pipeline}
& \multicolumn{3}{c}{\textbf{Semantic}}
& \multicolumn{6}{c}{\textbf{Cluster quality}}
& \multicolumn{3}{c}{\textbf{Graph quality}} \\
\cmidrule(lr){2-4}
\cmidrule(lr){5-10}
\cmidrule(lr){11-13}
& \multicolumn{3}{c}{\textbf{Coverage}}
& \multicolumn{3}{c}{\textbf{Silhouette}}
& \multicolumn{3}{c}{\textbf{ARI}}
& \multicolumn{3}{c}{\textbf{Modularity}} \\
\cmidrule(lr){2-4}
\cmidrule(lr){5-7}
\cmidrule(lr){8-10}
\cmidrule(lr){11-13}
& \textbf{Normal} & \textbf{Biblio.} & \textbf{No query}
& \textbf{Normal} & \textbf{Biblio.} & \textbf{No query}
& \textbf{Normal} & \textbf{Biblio.} & \textbf{No query}
& \textbf{Normal} & \textbf{Biblio.} & \textbf{No query} \\
\midrule
\midrule

\multicolumn{13}{l}{\textit{Mean rank}} \\
\midrule
Blind
& 5.85 & 6.20 & 6.15
& 5.10 & 4.55 & 5.25
& 5.55 & 5.50 & 5.80
& 5.40 & 4.80 & 5.35 \\

Corpus
& 6.30 & 5.60 & 5.80
& 4.10 & 4.15 & 3.80
& 4.90 & 5.00 & 4.65
& 3.70 & 4.10 & 3.95 \\

Corpus-select
& 3.55 & 2.80 & \textbf{2.35}
& 3.35 & 3.80 & 3.55
& 4.55 & 4.45 & 4.50
& \textbf{3.00} & 3.65 & 3.75 \\

Labeled
& \textbf{1.90} & \textbf{2.35} & 3.20
& 3.35 & \textbf{3.25} & \textbf{3.15}
& \textbf{2.25} & \textbf{2.00} & \textbf{2.50}
& 3.90 & 3.35 & 3.93 \\

Labeled-select
& 3.95 & 2.90 & 2.70
& 4.10 & 3.85 & 3.55
& 3.15 & 3.10 & 2.60
& 4.10 & 4.10 & 3.13 \\

Ranked
& 2.65 & 4.00 & 3.50
& \textbf{3.30} & 3.60 & 3.70
& 2.35 & 2.55 & \textbf{2.50}
& 3.10 & \textbf{3.20} & \textbf{3.00} \\

Human
& 3.80 & 4.15 & 4.30
& 4.70 & 4.80 & 5.00
& 5.25 & 5.40 & 5.45
& 4.80 & 4.80 & 4.90 \\

\midrule
\multicolumn{13}{l}{\textit{Median}} \\
\midrule
Blind
& 0.4085 & 0.4162 & 0.4157
& 0.0276 & 0.0336 & 0.0319
& 0.0496 & 0.0413 & 0.0320
& 0.1082 & 0.1203 & 0.1115 \\

Corpus
& 0.4021 & 0.4227 & 0.4228
& 0.0395 & 0.0338 & 0.0356
& 0.0700 & 0.0771 & 0.0594
& 0.1617 & 0.1637 & 0.1692 \\

Corpus-select
& 0.4234 & 0.4385 & \textbf{0.4452}
& 0.0369 & 0.0359 & 0.0425
& 0.0650 & 0.0635 & 0.0723
& 0.1422 & 0.1349 & 0.1501 \\

Labeled
& \textbf{0.4512} & \textbf{0.4439} & 0.4409
& \textbf{0.0436} & \textbf{0.0432} & \textbf{0.0504}
& \textbf{0.1661} & \textbf{0.1686} & \textbf{0.1525}
& \textbf{0.2071} & 0.2015 & 0.1830 \\

Labeled-select
& 0.4201 & 0.4300 & 0.4384
& 0.0402 & 0.0399 & 0.0485
& 0.1190 & 0.1106 & 0.1267
& 0.1928 & 0.1864 & \textbf{0.2247} \\

Ranked
& 0.4280 & 0.4345 & 0.4328
& 0.0421 & 0.0417 & 0.0472
& 0.1299 & 0.1344 & 0.1437
& 0.2059 & \textbf{0.2171} & 0.2119 \\

Human
& 0.4336 & 0.4336 & 0.4336
& 0.0319 & 0.0319 & 0.0319
& 0.0438 & 0.0438 & 0.0438
& 0.1113 & 0.1113 & 0.1113 \\

\midrule
\multicolumn{13}{l}{\textit{Win percentage}} \\
\midrule
Blind
& 0.00 & 0.00 & 5.00
& 0.00 & 0.00 & 10.00
& 5.00 & 0.00 & 0.00
& 0.00 & 15.00 & 0.00 \\

Corpus
& 0.00 & 0.00 & 0.00
& \textbf{20.00} & 25.00 & 15.00
& 0.00 & 0.00 & 10.00
& 15.00 & 5.00 & 15.00 \\

Corpus-select
& 5.00 & 30.00 & \textbf{30.00}
& 15.00 & 5.00 & 15.00
& 10.00 & 5.00 & 0.00
& \textbf{30.00} & 15.00 & 15.00 \\

Labeled
& \textbf{35.00} & 20.00 & 5.00
& 10.00 & \textbf{35.00} & \textbf{30.00}
& \textbf{35.00} & \textbf{45.00} & 25.00
& 10.00 & \textbf{35.00} & 10.00 \\

Labeled-select
& 10.00 & 10.00 & 20.00
& \textbf{20.00} & 5.00 & 10.00
& 20.00 & 15.00 & \textbf{30.00}
& 15.00 & 0.00 & 20.00 \\

Ranked
& 25.00 & 5.00 & 10.00
& \textbf{20.00} & 20.00 & 10.00
& 30.00 & 25.00 & 25.00
& 25.00 & 30.00 & \textbf{30.00} \\

Human
& 25.00 & \textbf{35.00} & \textbf{30.00}
& 15.00 & 10.00 & 10.00
& 0.00 & 10.00 & 10.00
& 5.00 & 5.00 & 10.00 \\

\bottomrule
\end{tabular}%
}
\end{table*}

\subsection{Evidence-Size Ablation}

The fourth ablation study evaluated whether the amount of reference evidence requested from or provided to the model affects the quality of the generated cluster descriptions. We compared three evidence-size settings: 10, 20, and 40 references per cluster. For pipelines 1--5, this value was expressed in the prompt as the approximate number of references \texttt{[\#]} that the model should use when enough relevant papers were available. For \textit{Ranked}, the value instead controlled the number of top-ranked papers from each cluster that were provided to the model as input evidence. This ablation tests whether more compact or broader cluster-level evidence leads to better bibliometric cluster descriptions.

Table~\ref{tab:variant-results} shows that changing the amount of reference evidence affects performance, but the overall pattern remains stable. The \textit{Blind} pipeline remains weak across most settings, while the strongest results continue to come from structured pipelines. \textit{Labeled} performs best for several core metrics, especially coverage and ARI, and remains strong across all three evidence-size settings. Increasing the evidence size benefits some pipelines, particularly \textit{Corpus-select}, which improves in coverage as the setting increases from 10 to 40, and \textit{Labeled-select}, which obtains strong modularity and ARI results at larger settings. However, larger evidence sets do not uniformly improve performance; for example, \textit{Ranked} is competitive across ARI and modularity but does not consistently improve as more papers are provided. Overall, the results suggest that the amount of evidence matters, but evidence structure and cluster labels remain more important than simply increasing the number of references.

\subsection{Human Evaluation of Structured Pipeline Outputs}\label{app:human-evaluation}

To complement the automatic metrics, we conducted a small-scale human evaluation of all pipelines. Two independent evaluators assessed outputs for five randomly selected benchmark papers, with five clusters per paper, yielding 25 cluster descriptions per pipeline. They compared descriptions by thematic coherence, specificity, faithfulness to evidence, and usefulness for cluster interpretation. Both evaluators agreed that, among the structured pipelines, \emph{Ranked} produced the strongest descriptions. \emph{Labeled} and \emph{Labeled-select} were generally similar in quality, with \emph{Labeled} slightly preferred overall. Human descriptions were rated below to the structured pipelines and above \emph{Blind}, which was penalized for hallucinated references.

\section{Code Availability}

All code required to reproduce the experiments, including the benchmark construction pipeline, prompting scripts, evaluation code, and analysis utilities, is provided in an anonymized repository: \url{https://anonymous.4open.science/r/How-Much-Structure-Do-LLMs-Need-EF56/}.

\section{Computational Cost}

All experiments were run on a single laptop and required approximately three days of wall-clock time. The total API cost was approximately 500 USD, consisting of about 400 USD in OpenAI usage and 100 USD in Gemini usage.

\section{Use of AI Assistants}

Beyond their use in the experimental pipelines, LLMs were also used as writing and research assistants during the preparation of the paper, including for editing, contrasting ideas, and supporting literature search.

\begin{table*}[t]
\centering
\caption{Benchmark results by experiment variant. Median summarizes central performance, mean rank summarizes relative ordering across instances, and win percentage reports how often each pipeline achieves the best score. Bold values indicate the best result for each metric and variant within each statistic.}
\label{tab:variant-results}
\small
\resizebox{\textwidth}{!}{%
\begin{tabular}{lcccccccccccc}
\toprule
\textbf{Pipeline}
& \multicolumn{3}{c}{\textbf{Semantic}}
& \multicolumn{6}{c}{\textbf{Cluster quality}}
& \multicolumn{3}{c}{\textbf{Graph quality}} \\
\cmidrule(lr){2-4}
\cmidrule(lr){5-10}
\cmidrule(lr){11-13}
& \multicolumn{3}{c}{\textbf{Coverage}}
& \multicolumn{3}{c}{\textbf{Silhouette}}
& \multicolumn{3}{c}{\textbf{ARI}}
& \multicolumn{3}{c}{\textbf{Modularity}} \\
\cmidrule(lr){2-4}
\cmidrule(lr){5-7}
\cmidrule(lr){8-10}
\cmidrule(lr){11-13}
& \textbf{10} & \textbf{20} & \textbf{40}
& \textbf{10} & \textbf{20} & \textbf{40}
& \textbf{10} & \textbf{20} & \textbf{40}
& \textbf{10} & \textbf{20} & \textbf{40} \\
\midrule

\multicolumn{13}{l}{\textit{Mean rank}} \\
\midrule
Blind
& 5.85 & 6.50 & 6.35
& 5.10 & 5.45 & 5.10
& 5.55 & 5.65 & 5.90
& 5.400 & 4.800 & 5.550 \\

Corpus
& 6.30 & 5.30 & 5.50
& 4.10 & 3.75 & \textbf{3.50}
& 4.90 & 4.75 & 4.90
& 3.700 & 3.850 & 3.700 \\

Corpus-select
& 3.55 & 2.65 & \textbf{2.35}
& 3.35 & 4.05 & 3.65
& 4.55 & 4.80 & 4.35
& \textbf{3.000} & 3.550 & 3.150 \\

Labeled
& \textbf{1.90} & \textbf{2.30} & 2.60
& 3.35 & 3.50 & 3.70
& \textbf{2.25} & \textbf{2.00} & \textbf{2.35}
& 3.900 & 3.775 & \textbf{2.975} \\

Labeled-select
& 3.95 & 2.85 & 2.55
& 4.10 & \textbf{3.20} & 3.65
& 3.15 & 2.60 & 2.55
& 4.100 & \textbf{3.250} & 3.625 \\

Ranked
& 2.65 & 4.25 & 4.25
& \textbf{3.30} & \textbf{3.20} & 3.95
& 2.35 & 2.35 & \textbf{2.35}
& 3.100 & 3.425 & 3.850 \\

Human
& 3.80 & 4.15 & 4.40
& 4.70 & 4.85 & 4.45
& 5.25 & 5.85 & 5.60
& 4.800 & 5.350 & 5.150 \\

\midrule
\multicolumn{13}{l}{\textit{Median}} \\
\midrule
Blind
& 0.4085 & 0.4159 & 0.4167
& 0.0276 & 0.0276 & 0.0247
& 0.0496 & 0.0474 & 0.0312
& 0.1082 & 0.1300 & 0.1364 \\

Corpus
& 0.4021 & 0.4271 & 0.4284
& 0.0395 & 0.0388 & 0.0400
& 0.0700 & 0.0778 & 0.0613
& 0.1617 & 0.1526 & 0.1716 \\

Corpus-select
& 0.4234 & 0.4453 & 0.4482
& 0.0369 & 0.0355 & 0.0388
& 0.0650 & 0.0788 & 0.0720
& 0.1422 & 0.1603 & 0.1914 \\

Labeled
& \textbf{0.4512} & \textbf{0.4469} & \textbf{0.4510}
& \textbf{0.0436} & 0.0439 & \textbf{0.0439}
& \textbf{0.1661} & 0.1439 & \textbf{0.1859}
& \textbf{0.2071} & 0.2060 & 0.2095 \\

Labeled-select
& 0.4201 & 0.4436 & 0.4457
& 0.0402 & \textbf{0.0465} & 0.0412
& 0.1190 & \textbf{0.1474} & 0.1342
& 0.1928 & \textbf{0.2159} & \textbf{0.2129} \\

Ranked
& 0.4280 & 0.4382 & 0.4360
& 0.0421 & 0.0442 & 0.0405
& 0.1299 & 0.1444 & 0.1761
& 0.2059 & 0.2015 & 0.1893 \\

Human
& 0.4336 & 0.4336 & 0.4335
& 0.0319 & 0.0319 & 0.0318
& 0.0438 & 0.0438 & 0.0437
& 0.1113 & 0.1065 & 0.1069 \\

\midrule
\multicolumn{13}{l}{\textit{Win percentage}} \\
\midrule
Blind
& 0.00 & 0.00 & 0.00
& 0.00 & 0.00 & 5.00
& 5.00 & 0.00 & 0.00
& 0.00 & 0.00 & 5.00 \\

Corpus
& 0.00 & 0.00 & 0.00
& \textbf{20.00} & 15.00 & 15.00
& 0.00 & 0.00 & 0.00
& 15.00 & 15.00 & 15.00 \\

Corpus-select
& 5.00 & 20.00 & \textbf{35.00}
& 15.00 & 10.00 & 15.00
& 10.00 & 0.00 & 5.00
& \textbf{30.00} & 20.00 & \textbf{30.00} \\

Labeled
& \textbf{35.00} & \textbf{35.00} & 25.00
& 10.00 & \textbf{25.00} & 5.00
& \textbf{35.00} & \textbf{45.00} & 20.00
& 10.00 & 20.00 & 25.00 \\

Labeled-select
& 10.00 & 10.00 & 10.00
& \textbf{20.00} & 20.00 & \textbf{30.00}
& 20.00 & 15.00 & \textbf{45.00}
& 15.00 & 15.00 & 10.00 \\

Ranked
& 25.00 & 0.00 & 5.00
& \textbf{20.00} & 20.00 & 15.00
& 30.00 & 40.00 & 30.00
& 25.00 & \textbf{25.00} & 10.00 \\

Human
& 25.00 & \textbf{35.00} & 25.00
& 15.00 & 10.00 & 15.00
& 0.00 & 0.00 & 0.00
& 5.00 & 5.00 & 5.00 \\

\bottomrule
\end{tabular}%
}
\end{table*}

\end{document}